\documentclass[journal]{IEEEtran}
\usepackage{setspace}
\usepackage[numbers]{natbib}
\usepackage{float}
\usepackage{subfig}
\usepackage{booktabs}
\usepackage{tabularx}
\usepackage{graphicx}
\usepackage{url}
\usepackage{xcolor}
\usepackage{stfloats}
\usepackage{comment}
\usepackage{amsmath}
\usepackage{amsfonts}
\usepackage{multirow}
\usepackage{booktabs}
\usepackage{listings}
\usepackage{pifont}
\usepackage{makecell}
\usepackage{nicematrix}
\usepackage{bm}
\usepackage{threeparttable}

\definecolor{verylightgray}{rgb}{.97,.97,.97}
\usepackage{algorithmicx}
\usepackage{fontawesome}
\usepackage[ruled,vlined,linesnumbered,boxed,commentsnumbered]{algorithm2e}
\usepackage{graphicx}
\usepackage{subcaption}
\usepackage[table]{xcolor}
\usepackage{afterpage}
\usepackage{hyperref}
\usepackage{etoolbox}
\AtBeginEnvironment{table}{\small}
\usepackage{amssymb}
\usepackage{pifont}
\usepackage{pifont}
\usepackage{graphicx}
\usepackage{xcolor}
\usepackage{tikz}
\newcommand{\badge}[2]{%
\tikz[baseline=(X.base)] 
\node[%
  draw=none,
  fill=#1,
  rounded corners=2pt,
  inner sep=2pt,
  minimum size=0.9em,
  text=white,
  font=\bfseries\small
](X){#2};
}

\ifCLASSINFOpdf
\else
\fi
\begin{document}
%
\title{Tactile-based Multimodal Fusion in Embodied Intelligence: A Survey of Vision, Language, and Contact-Driven Paradigms}
%
%
%

\author{Zhixiang Cao,
        Di Tian,
        Runwei Guan$^*$,
        Yanzhou Mu,
        Xiaolou Sun,
        Shaofeng Liang,
        Daizong Liu,
        Tao Huang,
        Yutao Yue,
        Henghui Ding,
        Bin Fang,
        Alex Zhou,
        Qing-Long Han,~\IEEEmembership{Fellow,~IEEE},
        and Hui Xiong$^*$,~\IEEEmembership{Fellow,~IEEE}

\thanks{Zhixiang Cao is with School of Electronic Science and Engineering, Xi'an Jiaotong University, China.}
\thanks{Runwei Guan, Di Tian, Shaofeng Liang, Yutao Yue and Hui Xiong are with Thrust of Artificial Intelligence, The Hong Kong University of Science and Technology (Guangzhou), China.}
\thanks{Yanzhou Mu is with State Key Laboratory for Novel Software Technology, Nanjing University, China.}
\thanks{Xiaolou Sun is with Purple Mountain Laboratory, China.}
\thanks{Daizong Liu is with Institute for Math \& AI, Wuhan University, China.}
\thanks{Tao Huang is with the Centre for AI and Data Science Innovation and the School of Science and Engineering, James Cook University, Australia.}
\thanks{Bin Fang is with School of Artificial Intelligence, Beijing University of Posts and Telecommunications, China.}
\thanks{Henghui Ding is with Institute of Big Data, Fudan University, China.}
\thanks{Alex Zhou is with Linkerbot (Beijing) Technology Co., Ltd, China.}
\thanks{Qing-Long Han is with School of Engineering, Swinburne University of Technology, Melbourne.}
\thanks{$^*$Corresponding author: \{runwayrwguan, xionghui\}@hkust-gz.edu.cn}}

%
%

\markboth{}%
{Shell \MakeLowercase{\textit{et al.}}: Bare Demo of IEEEtran.cls for IEEE Journals}
%



\maketitle

\begin{abstract}

Tactile sensing is a fundamental modality for embodied intelligence, offering unique and direct feedback on contact geometry, material properties, and interaction dynamics that remote sensors cannot replace. However, unimodal tactile perception is inherently limited by its sparse spatial coverage and lack of global semantic context. With the recent explosion in deep learning and large language models, integrating tactile with vision and language has become essential to bridge physical interaction with semantic reasoning, leading to the emergence of Multimodal Tactile Fusion. Despite rapid progress, the existing researches remain fragmented across disparate datasets, sensing modalities, and tasks, lacking a unified theoretical framework.
To address this gap, this paper provides a comprehensive survey of multimodal tactile fusion research up to the first quarter of 2026. We propose a hierarchical taxonomy that organizes the field into two primary dimensions: multimodal datasets and multimodal methods. On the data side, we categorize resources ranging from Tactile-Vision datasets, Tactile-Language datasets, Tactile-Vision-Language datasets, and Tactile-Vision-Other datasets. On the method side, we structure prior work into three core pillars: (1) Multimodal Perception and Recognition, which focuses on object understanding and grasp prediction; (2) Cross-Modal Generation, focusing on bidirectional translation between tactile, vision, and text; and (3) Multimodal Interaction, emphasizing feedback control and language-guided manipulation. Furthermore, we summarize representative tactile sensing hardware, review commonly used evaluation metrics and benchmark settings, and discuss current challenges and promising future directions. This survey aims to consolidate the current research landscape and provide a clear foundation for the development of more general, scalable, and robust tactile fusion systems. Paper collection is at \textbf{\href{https://github.com/Wayne-coding/Multimodal-Tactile-Fusion}{Wayne-coding/Multimodal-Tactile-Fusion}}.
\end{abstract}


\maketitle

%
\IEEEpeerreviewmaketitle

\section{Introduction}
\label{sec:intro}

Tactile perception is a cornerstone of embodied intelligence, bridging the gap between passive observation and active physical interaction. Unlike distal modalities such as vision, touch provides direct and proximal feedback on surface texture, material properties, and contact dynamics, which are critical for resolving visual ambiguity. For embodied agents, tactile feedback is not merely a supplementary modality but a fundamental component of the perception--action loop, enabling precise manipulation and stable grasping in contact-rich environments. The integration of multisensory cues, especially the synergy between vision and touch, is therefore essential for building robust perception and control systems that allow agents to operate effectively in the physical world.

Inspired by biological systems, tactile sensing has become an important component of robotic embodiment and computational intelligence. Recent studies~\cite{dave2024multimodal,yang2024binding,fu2024touch,cheng2024towards,ma2025cltp,cheng2025touch100k,zhang2025vtla,xie2025universal,feng2025anytouch} have increasingly explored the integration of tactile signals with vision~\cite{luo2018vitac}, language~\cite{ma2025cltp}, and other modalities such as audio or action streams~\cite{tong2025can}. This trend has advanced several important directions, including dexterous manipulation with closed-loop vision--tactile feedback~\cite{tong2025can}, physical property estimation of texture, compliance, and local geometry~\cite{kansana2025surformerv1,cho2025ra}, and embodied systems in which tactile experience provides physical grounding for semantic reasoning. These developments highlight the growing role of tactile information in connecting high-level perception with effective physical interaction.

Meanwhile, advances in Deep Learning (DL) have further accelerated multimodal tactile fusion. Beyond conventional architectures~\cite{dargan2020survey}, Reinforcement Learning (RL) and Transformer-based models~\cite{vaswani2017attention} have shown strong capability in modeling complex dependencies within high-dimensional sensory streams. More recently, Large Language Models (LLMs) and multimodal foundation models~\cite{zhao2023survey,girdhar2023imagebind} have pushed the field toward more unified representations, coupling tactile perception with semantic reasoning and action. These advances provide an important foundation for building general-purpose tactile systems~\cite{fu2024touch} that can operate in unstructured real-world environments.

\begin{figure*}[!h]
    \centering
    \includegraphics[width=0.9\textwidth]{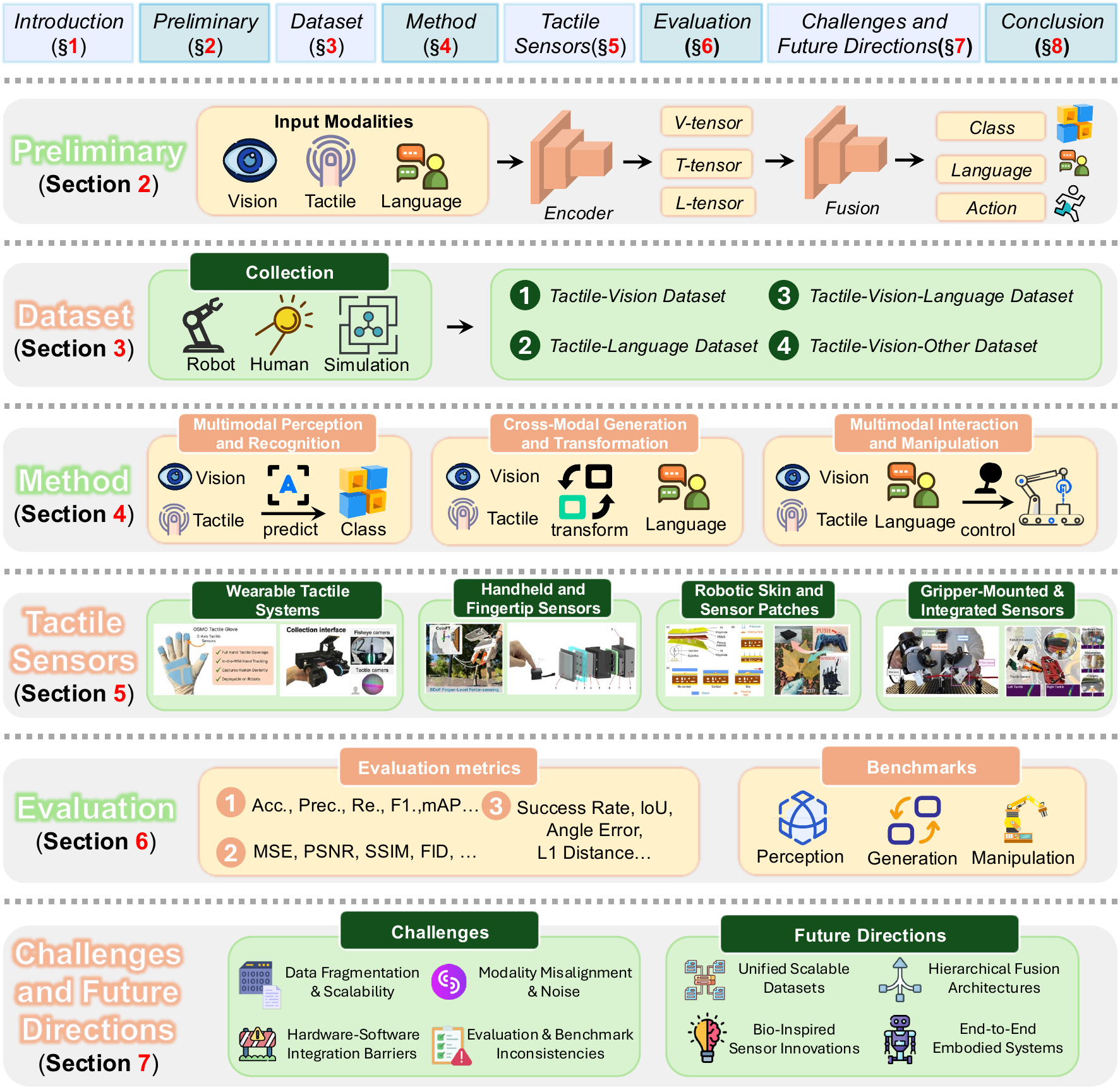}
     \caption{Overview of the structure of this survey on multimodal tactile fusion. 
}   
     \label{overview}
\end{figure*}

As illustrated in the hierarchical taxonomy (Fig.~\ref{overview}), this survey systematically organize multimodal tactile fusion into three primary pillars: Multimodal Datasets, Multimodal Methods, and Tactile Sensors. The dataset pillar provides a comprehensive overview of available resources, including Tactile-Vision datasets, Tactile-Language datasets, Tactile-Vision-Language datasets, and Tactile-Vision-Other datasets that incorporate additional modalities such as action, audio, or proprioception. The method pillar comprises into three core research categories: 
(1) Multimodal Perception and Recognition, covering tasks, such as material understanding, grasp-related prediction, and cross-modal retrieval; 
(2) Multimodal Cross-Modal Generation and Transformation, focusing on bidirectional generation and translation between tactile signals and other modalities; and 
(3) Multimodal Interaction and Manipulation, emphasizing contact-rich control and language-guided robotic manipulation. 
In parallel, the hardware pillar categorizes tactile sensing platforms into four groups according to their physical embodiment and deployment settings: wearable tactile systems, handheld and fingertip sensors, robotic skin and multimodal sensor patches, and gripper-mounted and integrated sensors. 
This structured perspective spans the full pipeline from hardware transduction to high-level reasoning, which enables us to examine current limitations and to outline future directions toward general-purpose tactile intelligence.

\begin{figure*}[htb]
    \centering
    \includegraphics[width=1\textwidth]{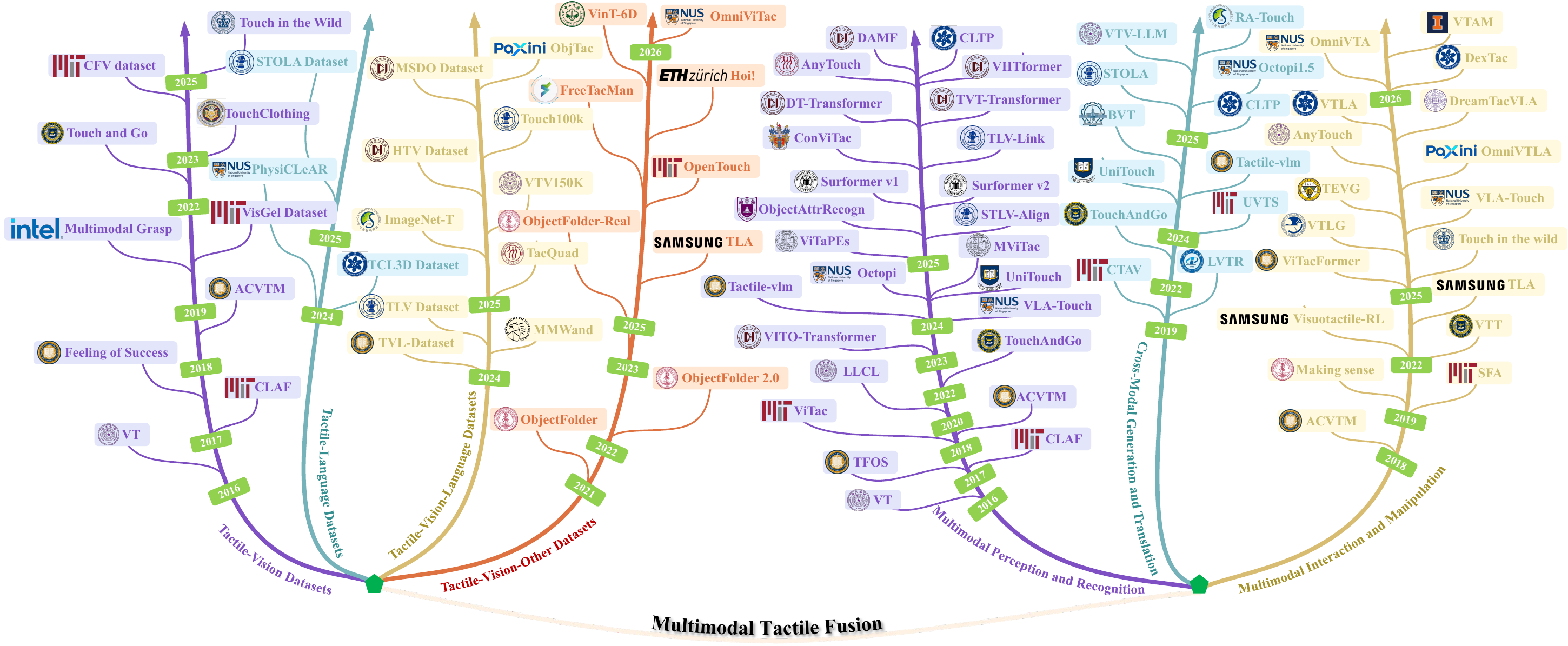}
     \caption{Overview of representative datasets, methods in multimodal tactile fusion.}
     \label{allpaper}
\end{figure*}

Fig.~\ref{allpaper} provides an overview of the representative datasets and methods discussed in this survey. By organizing the literature according to major taxonomy categories and chronological order, the figure highlights the developmental trajectory of multimodal tactile fusion and shows how the field has expanded across perception, generation, and interaction tasks. Compared with the existing surveys~\cite{gao2024transformer,sun2025tactile}, which mainly focus on specific modality pairs or model families, our taxonomy provides a broader system-level view of multimodal tactile fusion by linking data resources, learning methods, sensing hardware, and embodied interaction within a unified perspective. This formulation helps clarify the current research landscape and supports the subsequent discussion of open challenges and future research directions.

In summary, the contributions of this survey are as follows:
\begin{itemize}

    \item  
    To the best of the authors' knowledge, this is the first survey dedicated to multimodal tactile fusion, providing a unified perspective on how tactile signals are integrated into multimodal systems.

    \item 
    It introduces a hierarchical taxonomy that organizes multimodal tactile fusion from three complementary perspectives, namely multimodal datasets, multimodal methods, and tactile sensors, together with fine-grained subcategories.

    \item 
    It systematically reviews representative datasets, model architectures and applications developed up to the first quarter of 2026, and summarizes the major methodological developments and recent technical progress in the field.

    \item 
    It identifies current limitations, discuss es open challenges, and outline promising directions for advancing general-purpose multimodal tactile systems.

\end{itemize}

As shown in Fig.~\ref{overview}, the remainder of this survey is organized as follows. Section~\ref{sec:preliminary} introduces the preliminary foundations of multimodal tactile fusion, including input modalities, representation pipelines, and fusion mechanisms. Section~\ref{sec:datasets} reviews multimodal tactile datasets across different sensing modalities and data collection settings. Section~\ref{sec:methods} presents a systematic taxonomy and analysis of multimodal tactile fusion methods. Section~\ref{sec:tactile_sensors} surveys tactile sensing technologies and sensor designs that underpin data acquisition and physical interaction. Section~\ref{sec:evaluation} summarizes evaluation metrics and benchmarks used across perception, generation, and manipulation tasks. Section~\ref{sec:discussion} discusses open challenges and future research directions. Finally, Section~\ref{sec:Conclusion} concludes the survey.

\section{Preliminary}
\label{sec:preliminary}

\subsection{Problem Scope and Definition}

Multimodal tactile fusion~\cite{dave2024multimodal, yang2024binding, fu2024touch, cheng2024towards, ma2025cltp, cheng2025touch100k, zhang2025vtla, xie2025universal, feng2025anytouch} examines how tactile information is integrated with other modalities to support perception, generation, and interaction in robotic systems. In this survey, we focus on settings in which tactile sensing plays a central role and combines with vision and/or language. Specifically, we consider the bimodal settings T-V and T-L, as well as the trimodal setting T-V-L. Recent studies~\cite{yang2024binding, fu2024touch, cheng2024towards, ma2025cltp, cheng2025touch100k, zhang2025vtla} have increasingly adopted the T-V-L setting because it provides a unified framework for connecting physical interaction, visual context, and semantic reasoning.

From a modeling perspective, multimodal tactile fusion can be viewed as a multi-stage pipeline that maps physical interaction signals to computational representations. Unlike vision or language, tactile signals are inherently contact-driven and arise only through physical interaction. Accordingly, the multimodal tactile fusion process comprises the following hierarchical stages.

\subsubsection{Physical Transduction and Spatiotemporal Observation}
During interaction, tactile sensors convert physical stimuli such as deformation, force, or vibration into digital signals. The raw inputs can be represented as
\begin{equation}
    x_t \in \mathbb{R}^{S_t \times T_t \times C_t},\quad x_v \in \mathbb{R}^{S_v \times T_v \times C_v},\quad x_l \in \mathcal{V}^{L},
\end{equation}
where $x_t$ denotes tactile observations with spatial configuration $S_t$, temporal length $T_t$, and channel dimension $C_t$, while $x_v$ and $x_l$ denote visual and language inputs, respectively.

\subsubsection{Modality-Specific Representation Learning}
Because different modalities have heterogeneous input formats, each modality is encoded into a latent representation:
\begin{equation}
    z_m = \mathcal{E}_m(x_m; \theta_m), \quad m \in \{t, v, l\},
\end{equation}
where $\mathcal{E}_m(\cdot)$ denotes a modality-specific encoder. This step transforms tactile, visual, and language inputs into learned representations, yielding $z_t, z_v, z_l \in \mathbb{R}^d$ when all three modalities are available.

\subsubsection{Cross-Modal Fusion and Joint Representation}
The resulting unimodal representations are then integrated into a shared joint representation for downstream learning:
\begin{equation}
    z_{\mathrm{joint}} = \Phi(\{z_m\}_{m \in \mathcal{M}}; \theta_{\Phi}),
\end{equation}
where $\mathcal{M} \subseteq \{t, v, l\}$ denotes the set of available modalities, and $\Phi(\cdot)$ denotes the fusion operator. Depending on the method, this operator may be implemented through feature concatenation, cross-attention, or alignment-based learning.

\subsubsection{Embodied Decoding and Task Execution}
Finally, the fused representation is decoded for downstream tasks:
\begin{equation}
    y = \mathcal{D}(z_{\mathrm{joint}}; \theta_{\mathcal{D}}),
\end{equation}
where $y$ may denote a perception output, a generated cross-modal signal, or an action output for robotic control.

Within this framework, multimodal tactile fusion aims to learn complementary representations that overcome the limitations of single-modality perception and support more robust embodied decision-making.

\subsection{Conceptual Background}

\subsubsection{Modalities and Encoders}

Multimodal tactile fusion mainly involves tactile, visual, and language signals. Tactile inputs encode local geometry, texture, force, and contact dynamics, while visual inputs provide global shape and scene context. Language adds semantic descriptions of object properties, materials, and actions, linking tactile observations to high-level concepts or instructions. To integrate these heterogeneous modalities, recent methods usually adopt modality-specific encoders. Tactile and visual streams often use CNN- or Transformer-based backbones, such as ResNet~\cite{he2016deep} and ViT~\cite{dosovitskiy2020image}, whereas language is commonly modeled with pre-trained text encoders such as BERT~\cite{devlin2019bert} or OpenCLIP~\cite{radford2021learning}. In many recent studies, the vision and language branches are initialized with pre-trained models to provide stronger semantic priors and improve cross-modal alignment.

\subsubsection{Tactile Sensor Taxonomy}

From an engineering perspective, tactile sensors can be characterized along three dimensions~\cite{yu2021recent,li2025classification,meribout2024tactile,wan2017recent}. Structurally, they include single-point, array-based, and distributed designs. Mechanistically, piezoresistive, capacitive, piezoelectric, optical, and electromagnetic principles offer different trade-offs in sensitivity, spatial resolution, and dynamic response. In addition, substrate design---including rigid, flexible, or hybrid forms---affects the balance between sensing precision and physical compliance. These hardware differences directly influence the quality of tactile signals used in downstream multimodal fusion.

\subsubsection{Tactile Signal Acquisition and Representation}

Despite their hardware diversity, tactile sensors generally follow a common transduction pipeline~\cite{li2020review,yi2018biomimetic}. Contact-induced physical stimuli, such as pressure, deformation, or vibration, are converted into measurable electrical or optical signals. For example, vision-based tactile sensors capture surface deformations as image-like observations, making them naturally compatible with standard visual backbones. After basic signal conditioning, including amplification, filtering, and digitization, these raw measurements are mapped into task-specific latent representations through handcrafted features or end-to-end neural encoders. This process highlights that tactile observations, unlike distal vision or language, are inherently contact-driven and physically grounded, and their representations are strongly affected by both sensor morphology and interaction dynamics.

\subsubsection{Representative Mainstream Tactile Sensors}

\begin{figure*}[t]
    \centering
    \includegraphics[width=1\textwidth]{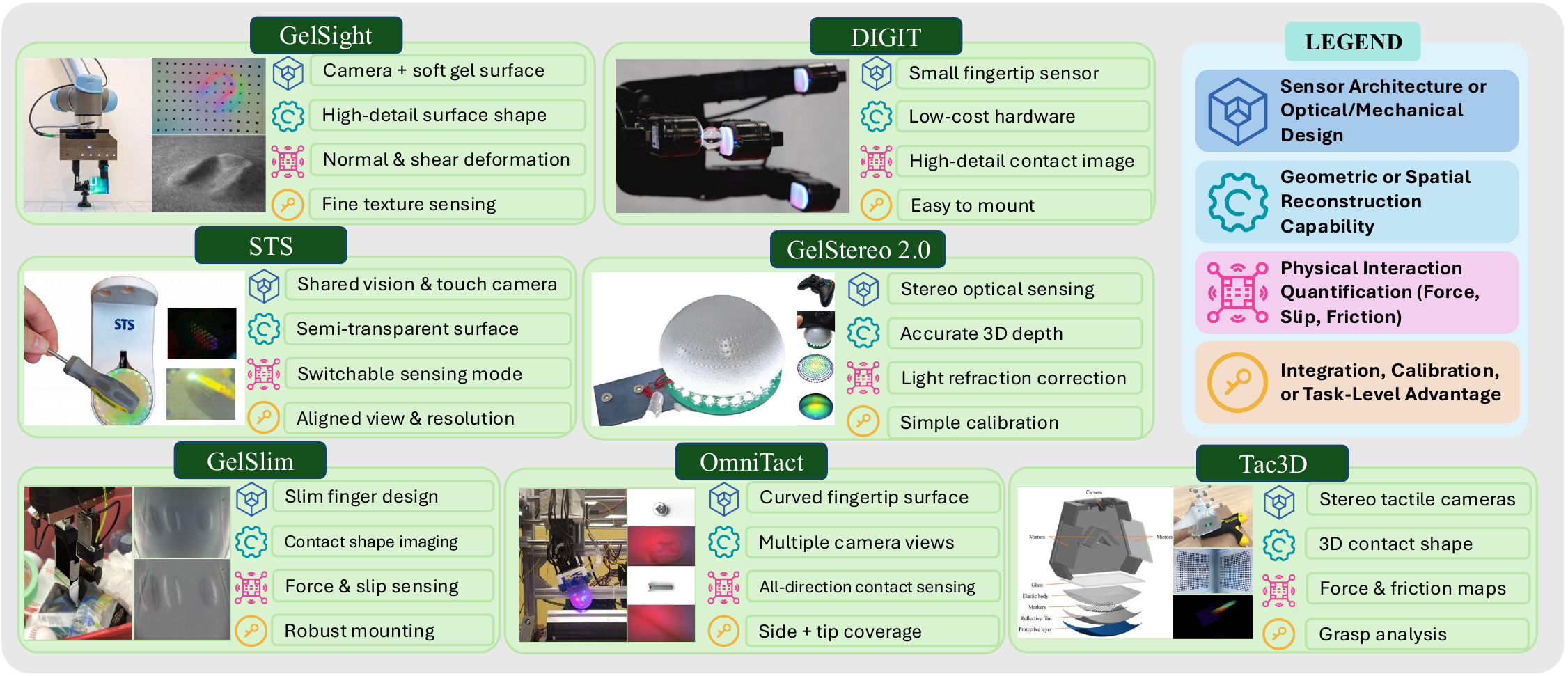}
    \caption{Representative tactile sensors.}
    \label{tactile_sensors}
\end{figure*}

Fig.~\ref{tactile_sensors} compares representative tactile sensors along four dimensions: sensor architecture, geometric reconstruction capability, physical interaction quantification, and integration or task-level advantages. Here, we briefly summarize the core characteristics of several widely used platforms.

\paragraph{GelSight~\cite{yuan2017gelsight}}
GelSight is a representative vision-based tactile sensor that uses a camera and a soft gel surface to capture high-resolution contact deformations. It is particularly effective for fine surface reconstruction and detailed sensing of normal and shear cues.

\paragraph{DIGIT~\cite{lambeta2020digit}}
DIGIT is a compact fingertip tactile sensor designed for low-cost deployment and practical integration. Its main advantage lies in preserving detailed contact imaging within a small and lightweight form factor.

\paragraph{Tac3D~\cite{zhang2022tac3d}}
Tac3D emphasizes stereo tactile sensing and 3D contact reconstruction. It also supports force and friction estimation, making it useful for grasp analysis.

\paragraph{GelStereo~2.0~\cite{zhang2023gelstereo}}
GelStereo~2.0 focuses on accurate stereo tactile perception. Its main strengths include reliable 3D depth estimation, refraction-aware reconstruction, and simplified calibration.

\paragraph{GelSlim~\cite{taylor2022gelslim}}
GelSlim is a slim tactile sensor for robotic fingers. It combines contact shape imaging with force and slip sensing, while remaining easy to mount on grippers.

\paragraph{OmniTact~\cite{padmanabha2020omnitact}}
OmniTact is designed for curved fingertip sensing with multiple internal views. This structure enables omnidirectional contact perception and coverage.

\paragraph{STS~\cite{hogan2021seeing}}
STS integrates visual and tactile sensing within a shared optical setup. Its main advantage is that it supports spatially aligned visual and tactile perception through a switchable sensing mode.

\subsubsection{Fusion, Alignment, and Training Paradigms}

A key challenge in multimodal tactile fusion~\cite{li2024multimodal} lies in effectively combining heterogeneous tactile, visual, and language representations, and optimizing the resulting models effectively. Existing methods~\cite{dave2024multimodal, yang2024binding, fu2024touch, cheng2024towards, ma2025cltp, cheng2025touch100k, zhang2025vtla, xie2025universal, feng2025anytouch, yin2024survey, li2025vhtformer} differ not only in where and how modalities are fused, but also in how cross-modal alignment is enforced and how the models are trained.

\paragraph{Early fusion and late fusion}
Early fusion combines modalities at the input level or shallow feature level, which encourages joint representation learning but is often more sensitive to noise, modality imbalance, or missing inputs. By contrast, late fusion merges modality-specific representations after separate encoding, thereby preserving modality-specific structure and offering greater flexibility for downstream decision-making.

\paragraph{Shared embedding spaces, cross-attention, and contrastive alignment}
Many methods map tactile, visual, and language features into a common embedding space to support cross-modal matching, retrieval, or representation alignment. More recent models further employ cross-attention to capture fine-grained interactions across tactile patches, visual regions, and language tokens. In addition, contrastive learning is widely used to pull paired samples closer and push unpaired samples apart, providing an effective objective for T-V, T-L, and T-V-L representation learning.

\paragraph{Supervised learning and transfer-based adaptation}
Supervised learning remains a common strategy for tasks with explicit labels, such as recognition and classification, but it is often constrained by the high cost of tactile data collection and annotation. To reduce this dependence, many methods adopt pre-training and transfer learning, where vision and language branches are initialized from large pre-trained models and then adapted to tactile tasks through supervised fine-tuning or task-specific adaptation.

\paragraph{Curriculum learning, reinforcement learning, and hybrid strategies}
For contact-rich interaction tasks, curriculum learning can improve optimization stability by gradually increasing task difficulty, while reinforcement learning is widely used when dense supervision is unavailable and tactile feedback is important for policy learning. In practice, many recent systems adopt hybrid strategies that combine pre-training, supervised fine-tuning, and reinforcement learning to balance generalization, data efficiency, and control performance.

\begin{figure}[htb]
    \centering
    \includegraphics[width=0.5\textwidth]{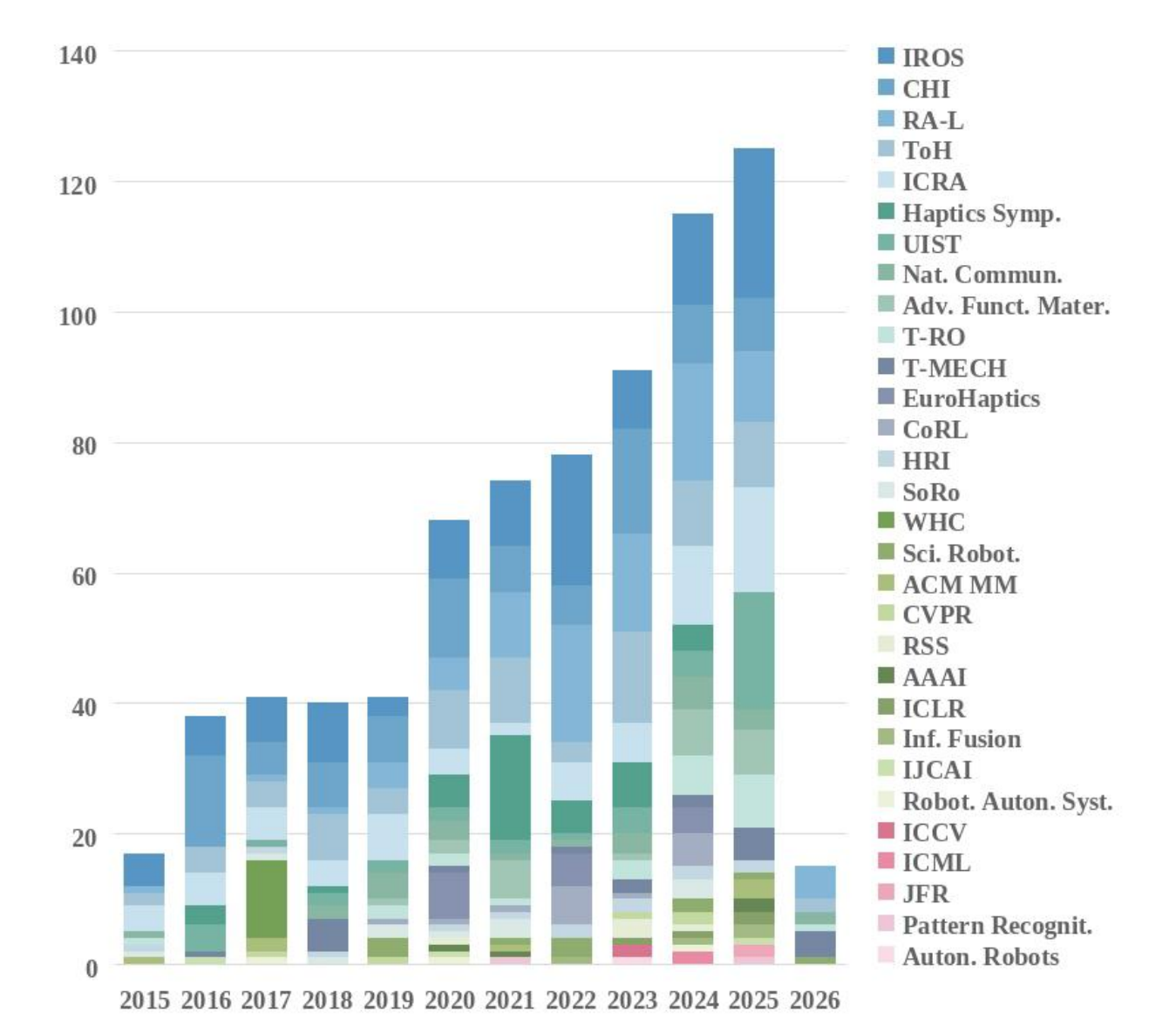}
     \caption{Publication trend of multimodal tactile fusion papers from 2015 to 2026.}
     \label{fig:pub_trend}
\end{figure}

\subsubsection{Survey Methodology and Publication Trend}
We survey multimodal tactile fusion studies by collecting papers from major journals and conferences in robotics, computer vision, haptics, and multimodal learning, and selecting works that combine tactile sensing with at least one additional modality under data-driven or learning-based settings. As shown in Fig.~\ref{fig:pub_trend}, the number of relevant publications has increased steadily over the past decade, with a clear acceleration after 2020. Early studies were mainly reported in robotics and haptics venues, whereas more recent works have spread to broader vision and multimodal communities, reflecting the growing recognition of tactile sensing as an essential modality for multimodal perception, interaction, and embodied intelligence.
\section{Multimodal Datasets}
\label{sec:datasets}

Datasets play a central role in multimodal tactile fusion because they pair tactile observations with complementary modalities, enabling the learning of cross-modal correspondence, semantic grounding, and interaction-aware representations that cannot be obtained from touch alone.

In this survey, we organize existing datasets into four groups according to their modality composition: T-V, T-L, T-V-L, and T-V-O, where ``O'' includes signals such as action, audio, and proprioception. This modality-based taxonomy reflects the evolution of the field from early bimodal benchmarks to more recent large-scale and semantically richer datasets. In the following, we review representative datasets in each group and summarize their sensing setups, scale, and supported tasks.

\begin{table*}[!ht]

\centering
\caption{Comparative summary for the datasets of tactile-based multimodal fusion.}

\resizebox{0.90\textwidth}{!}{
\begin{tabular}{l l c c c l l c c l}
\hline
\textbf{Category} &
\textbf{Dataset} &
\textbf{Year} &
\textbf{Object inst.} &
\textbf{Scale} &
\textbf{Modalities} &
\textbf{Source} &
\textbf{Real} &
\textbf{Environment} &
\textbf{Sensor Type} \\
\hline

\multirow{10}{*}{\textbf{T-V}} 

& VT~\cite{liu2016visual} &
2016&
18&
-& 
\includegraphics[height=0.9em]{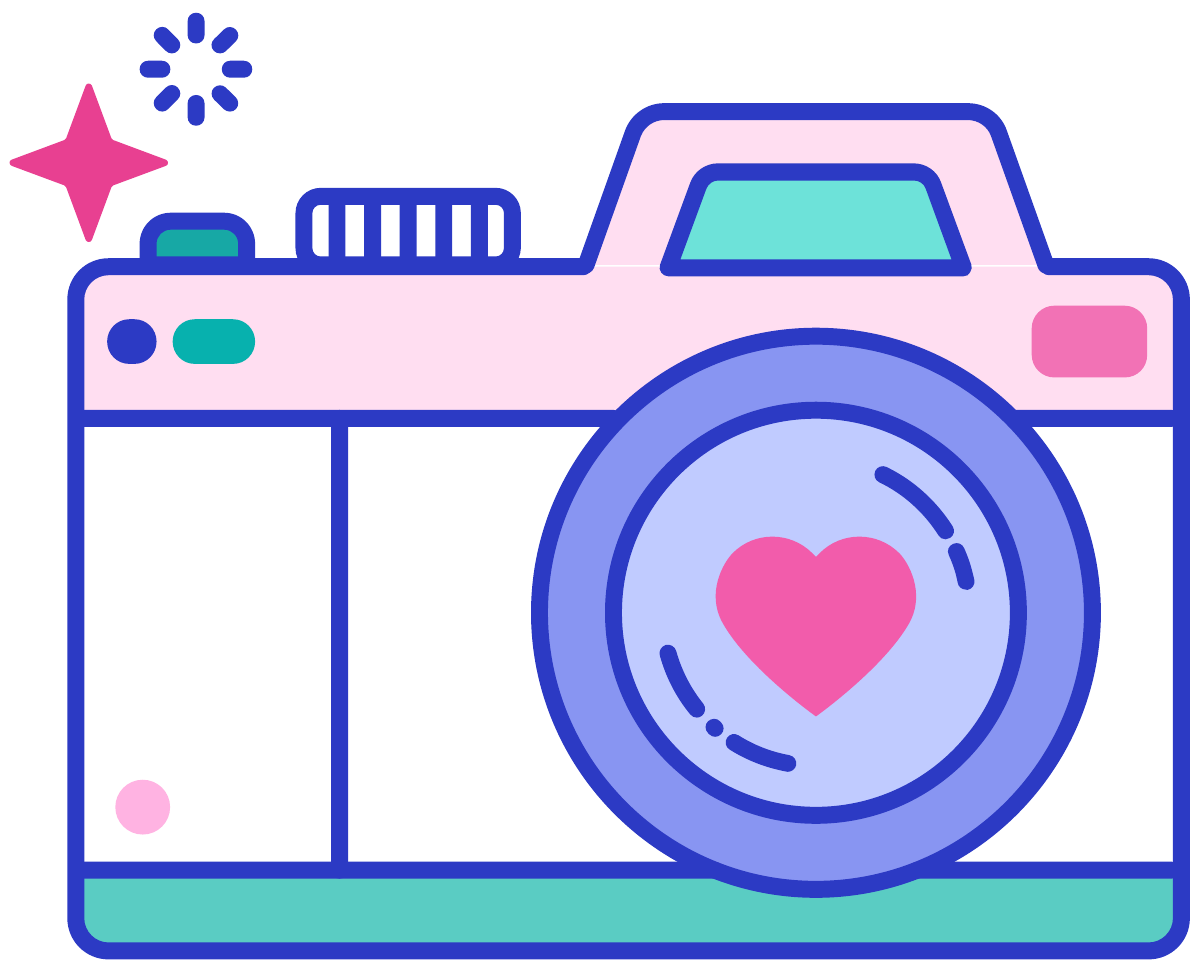} \includegraphics[height=0.9em]{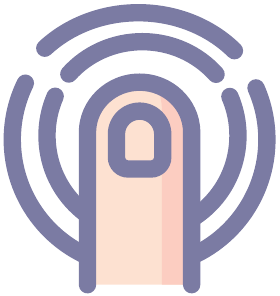} &
\includegraphics[height=0.9em]{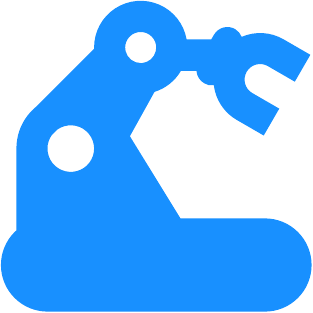}  &
\checkmark &
Indoor &
DIY \\

& Feeling of Success~\cite{calandra2017feeling} &
2017&
106 &
9.3k & 
\includegraphics[height=0.9em]{logo/vision_logo.pdf} \includegraphics[height=0.9em]{logo/touch_logo.pdf} &
\includegraphics[height=0.9em]{logo/Robot_logo.pdf}  &
\checkmark &
Lab &
\badge{orange}{G} \\

& CLAF~\cite{yuan2017connecting} &
2017&
118 &
2950 &
\includegraphics[height=0.9em]{logo/vision_logo.pdf} \includegraphics[height=0.9em]{logo/touch_logo.pdf}  &
\includegraphics[height=0.9em]{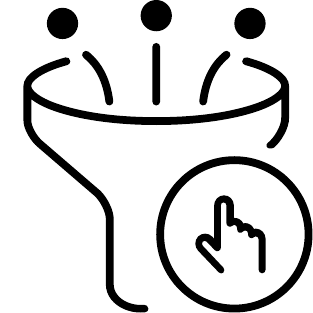} &
\checkmark &
Indoor  &
\badge{orange}{G}
\\

& ACVTM~\cite{calandra2018more} &
2018&
65 &
18070 & 
\includegraphics[height=0.9em]{logo/vision_logo.pdf} \includegraphics[height=0.9em]{logo/touch_logo.pdf}  &
\includegraphics[height=0.9em]{logo/Robot_logo.pdf} &
\checkmark &
Indoor  &
\badge{orange}{G}
\\

& Multimodal Grasp~\cite{wang2019multimodal} &
2019 &
10 &
2550 & 
\includegraphics[height=0.9em]{logo/vision_logo.pdf} \includegraphics[height=0.9em]{logo/touch_logo.pdf} &
\includegraphics[height=0.9em]{logo/Robot_logo.pdf}  &
\checkmark &
Lab &
\badge{orange}{G} \\

& VisGel~\cite{li2019connecting} &
2019 &
195 &
3M & 
\includegraphics[height=0.9em]{logo/vision_logo.pdf} \includegraphics[height=0.9em]{logo/touch_logo.pdf} &
\includegraphics[height=0.9em]{logo/Robot_logo.pdf}  &
\checkmark &
Indoor + Lab &
\badge{orange}{G} \\

& Touch and Go~\cite{yang2022touch} &
2022&
3971&
13.9k& 
\includegraphics[height=0.9em]{logo/vision_logo.pdf} \includegraphics[height=0.9em]{logo/touch_logo.pdf} &
\includegraphics[height=0.9em]{logo/Human_logo.pdf} &
\checkmark &
Indoor + Outdoor &
\badge{orange}{G} \\

& TouchClothing~\cite{gao2023controllable} &
2023&
20 &
4k& 
\includegraphics[height=0.9em]{logo/vision_logo.pdf} \includegraphics[height=0.9em]{logo/touch_logo.pdf} &
\includegraphics[height=0.9em]{logo/Human_logo.pdf} &
\checkmark &
Indoor &
\badge{orange}{G} \\

& CFV~\cite{li2023learning} &
2023&
89 &
150k & 
\includegraphics[height=0.9em]{logo/vision_logo.pdf} \includegraphics[height=0.9em]{logo/touch_logo.pdf} &
\includegraphics[height=0.9em]{logo/Human_logo.pdf} &
\checkmark &
Indoor &
DIY \\

& Touch in the Wild~\cite{zhu2025touch} &
2025&
-&
2.6m & 
\includegraphics[height=0.9em]{logo/vision_logo.pdf} \includegraphics[height=0.9em]{logo/touch_logo.pdf} &
\includegraphics[height=0.9em]{logo/Robot_logo.pdf}  &
\checkmark &
Indoor + Outdoor &
DIY \\

\hline

\multirow{4}{*}{\textbf{T-L}} 

& PhysiCLeAR~\cite{yu2024octopi} &
2024&
74 &
408 &
 \includegraphics[height=0.9em]{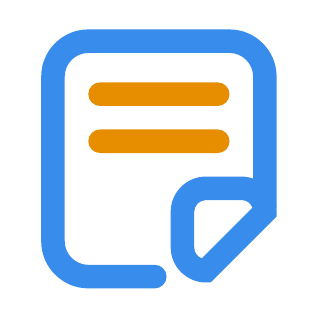} \includegraphics[height=0.9em]{logo/touch_logo.pdf} &
\includegraphics[height=0.9em]{logo/Robot_logo.pdf} \includegraphics[height=0.9em]{logo/Human_logo.pdf} &
\checkmark &
Lab &
\badge{orange}{G} \\

& TCL3D~\cite{ma2025cltp} &
2025&
117 &
50k &
 \includegraphics[height=0.9em]{logo/language_logo.pdf} \includegraphics[height=0.9em]{logo/touch_logo.pdf} &
\includegraphics[height=0.9em]{logo/Robot_logo.pdf} \includegraphics[height=0.9em]{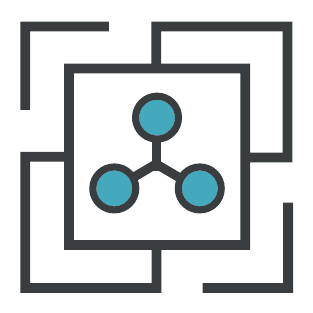} &
\checkmark &
Lab + Virtual &
\badge{gray}{D} \\

& STOLA~\cite{cheng2025stola} &
2025&
14 &
- &
\includegraphics[height=0.9em]{logo/language_logo.pdf} \includegraphics[height=0.9em]{logo/touch_logo.pdf}&
\includegraphics[height=0.9em]{logo/Human_logo.pdf} &
\checkmark &
xx &
xx \\

\hline

\multirow{10}{*}{\textbf{T-V-L}} 

& TVL Dataset~\cite{fu2024touch} &
2024&
- &
44k &
\includegraphics[height=0.9em]{logo/vision_logo.pdf} \includegraphics[height=0.9em]{logo/touch_logo.pdf} \includegraphics[height=0.9em]{logo/language_logo.pdf} &
\includegraphics[height=0.9em]{logo/Human_logo.pdf} &
\checkmark &
Lab + Outdoor &
\badge{gray}{D} \\

& MMWand~\cite{chi2024multi} &
2024&
102 &
5.2k &
\includegraphics[height=0.9em]{logo/vision_logo.pdf} \includegraphics[height=0.9em]{logo/touch_logo.pdf} \includegraphics[height=0.9em]{logo/language_logo.pdf} &
\includegraphics[height=0.9em]{logo/Human_logo.pdf} &
\checkmark &
Lab + Indoor &
\badge{orange}{G} \\

& TLV~\cite{cheng2024towards} &
2024&
- &
20k &
\includegraphics[height=0.9em]{logo/vision_logo.pdf} \includegraphics[height=0.9em]{logo/touch_logo.pdf} \includegraphics[height=0.9em]{logo/language_logo.pdf} &
\includegraphics[height=0.9em]{logo/Human_logo.pdf} &
\checkmark &
Indoor&
\badge{gray}{D}  \\

& TacQuad~\cite{feng2025anytouch} &
2025 &
124 &
72k & 
\includegraphics[height=0.9em]{logo/vision_logo.pdf} \includegraphics[height=0.9em]{logo/touch_logo.pdf} \includegraphics[height=0.9em]{logo/language_logo.pdf} &
\includegraphics[height=0.9em]{logo/Robot_logo.pdf} \includegraphics[height=0.9em]{logo/Human_logo.pdf} &
\checkmark &
Indoor + Outdoor &
\badge{orange}{G}
\badge{gray}{D}
\badge{teal}{T}\\

& ImageNet-T\cite{cho2025ra} &
2025&
-&
150k & 
\includegraphics[height=0.9em]{logo/vision_logo.pdf} \includegraphics[height=0.9em]{logo/touch_logo.pdf} \includegraphics[height=0.9em]{logo/language_logo.pdf}&
\includegraphics[height=0.9em]{logo/Human_logo.pdf} &
\checkmark &
Web images &
None \\

& Touch100k~\cite{cheng2025touch100k} &
2025&
4166 &
100k &
\includegraphics[height=0.9em]{logo/vision_logo.pdf} \includegraphics[height=0.9em]{logo/touch_logo.pdf} \includegraphics[height=0.9em]{logo/language_logo.pdf} &
\includegraphics[height=0.9em]{logo/Robot_logo.pdf} \includegraphics[height=0.9em]{logo/Human_logo.pdf}&
\checkmark &
Lab &
\badge{orange}{G} \\

& HTV~\cite{wang2025damf} &
2025&
17  &
850 &
\includegraphics[height=0.9em]{logo/vision_logo.pdf} \includegraphics[height=0.9em]{logo/touch_logo.pdf} \includegraphics[height=0.9em]{logo/language_logo.pdf} &
\includegraphics[height=0.9em]{logo/Robot_logo.pdf}  &
\checkmark &
Lab &
\badge{orange}{G} \\

& VTV150K~\cite{xie2025universal} &
2025&
100 &
150k &
\includegraphics[height=0.9em]{logo/vision_logo.pdf} \includegraphics[height=0.9em]{logo/touch_logo.pdf} \includegraphics[height=0.9em]{logo/language_logo.pdf} &
\includegraphics[height=0.9em]{logo/Robot_logo.pdf}  &
\checkmark &
Lab &
\badge{orange}{G}
\badge{gray}{D}
\badge{teal}{T} \\

& MSDO~\cite{li2025tvt} &
2025&
18 &
1800 &
\includegraphics[height=0.9em]{logo/vision_logo.pdf} \includegraphics[height=0.9em]{logo/touch_logo.pdf} \includegraphics[height=0.9em]{logo/language_logo.pdf} &
\includegraphics[height=0.9em]{logo/Robot_logo.pdf} \includegraphics[height=0.9em]{logo/Human_logo.pdf} &
\checkmark &
Lab &
\badge{gray}{D} \\

& ObjTac~\cite{cheng2025omnivtla} &
2025&
56 &
135k &
\includegraphics[height=0.9em]{logo/vision_logo.pdf} \includegraphics[height=0.9em]{logo/touch_logo.pdf} \includegraphics[height=0.9em]{logo/language_logo.pdf} &
\includegraphics[height=0.9em]{logo/Human_logo.pdf} &
\checkmark &
Lab &
\badge{gray}{D} \\

\hline

\multirow{3}{*}{\textbf{T-V-O}} 

& ObjectFolder~\cite{gao2021objectfolder} &
2021&
100 &
- & 
\includegraphics[height=0.9em]{logo/vision_logo.pdf} \includegraphics[height=0.9em]{logo/touch_logo.pdf} \includegraphics[height=0.9em]{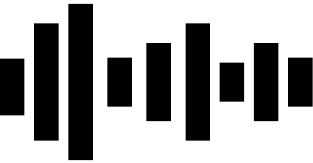} &
\includegraphics[height=0.9em]{logo/Simulation_logo.pdf} &
\ding{55} &
Virtual &
\badge{blue}{S}\\

& ObjectFolder 2.0~\cite{gao2022objectfolder} &
2022&
1000&
- & 
\includegraphics[height=0.9em]{logo/vision_logo.pdf} \includegraphics[height=0.9em]{logo/touch_logo.pdf} \includegraphics[height=0.9em]{logo/audio_logo.pdf} &
\includegraphics[height=0.9em]{logo/Simulation_logo.pdf} &
\ding{55} &
Virtual &
\badge{blue}{S}\\

& ObjectFolder-Real~\cite{gao2023objectfolder} &
2023&
100&
- & 
\includegraphics[height=0.9em]{logo/vision_logo.pdf} \includegraphics[height=0.9em]{logo/touch_logo.pdf} \includegraphics[height=0.9em]{logo/audio_logo.pdf} &
\includegraphics[height=0.9em]{logo/Robot_logo.pdf} \includegraphics[height=0.9em]{logo/Human_logo.pdf} &
\checkmark &
Indoor &
\badge{orange}{G} \\

& TLA~\cite{hao2025tla} &
2025&
- &
24k &
 \includegraphics[height=0.9em]{logo/language_logo.pdf} \includegraphics[height=0.9em]{logo/touch_logo.pdf} \includegraphics[height=0.9em]{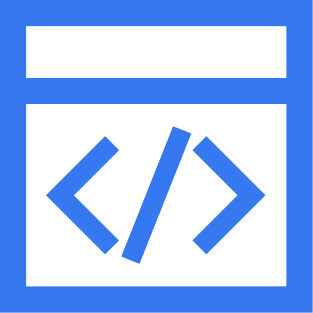} &
\includegraphics[height=0.9em]{logo/Robot_logo.pdf}  & 
\checkmark &
Lab &
\badge{pink}{G} \\

& FreeTacMan~\cite{wu2025freetacman} &
2025&
50 &
3m &
 \includegraphics[height=0.9em]{logo/vision_logo.pdf} \includegraphics[height=0.9em]{logo/touch_logo.pdf} \includegraphics[height=0.9em]{logo/action_logo.pdf} &
\includegraphics[height=0.9em]{logo/Robot_logo.pdf}  & 
\checkmark &
Lab &
DIY \\

& OpenTouch~\cite{song2025opentouch} &
2025&
8000&
- &
 \includegraphics[height=0.9em]{logo/vision_logo.pdf} \includegraphics[height=0.9em]{logo/language_logo.pdf} \includegraphics[height=0.9em]{logo/touch_logo.pdf} \includegraphics[height=0.9em]{logo/action_logo.pdf} &
\includegraphics[height=0.9em]{logo/Human_logo.pdf}  & 
\checkmark &
Outdoor &
DIY \\

& Hoi!~\cite{engelbracht2025hoi} &
2025&
381&
- &
\includegraphics[height=0.9em]{logo/vision_logo.pdf} \includegraphics[height=0.9em]{logo/touch_logo.pdf} \includegraphics[height=0.9em]{logo/action_logo.pdf} &
\includegraphics[height=0.9em]{logo/Human_logo.pdf}  & 
\checkmark &
Indoor &
\badge{gray}{D} \\

& VinT-6D~\cite{wan2024vint} &
2025&
10 &
2.1m &
\includegraphics[height=0.9em]{logo/vision_logo.pdf} \includegraphics[height=0.9em]{logo/touch_logo.pdf} \includegraphics[height=0.9em]{logo/action_logo.pdf} &
\includegraphics[height=0.9em]{logo/Robot_logo.pdf}  & 
\checkmark &
Indoor &
DIY \\

& OmniViTac~\cite{zheng2026omnivtavisuotactileworldmodeling} &
2026&
126 &
- &
\includegraphics[height=0.9em]{logo/vision_logo.pdf} \includegraphics[height=0.9em]{logo/touch_logo.pdf} \includegraphics[height=0.9em]{logo/action_logo.pdf} &
\includegraphics[height=0.9em]{logo/Human_logo.pdf}  & 
\checkmark &
Indoor &
DIY \\


\hline

\end{tabular}
}

\caption*{
\textbf{Input:} 
\includegraphics[height=0.9em]{logo/audio_logo.pdf} audio,
\includegraphics[height=0.9em]{logo/touch_logo.pdf} touch,
\includegraphics[height=0.9em]{logo/action_logo.pdf} action,
\includegraphics[height=0.9em]{logo/language_logo.pdf} language,
\includegraphics[height=0.9em]{logo/vision_logo.pdf} vision. \;
\textbf{Source:} 
\includegraphics[height=0.9em]{logo/Human_logo.pdf} Human,
\includegraphics[height=0.9em]{logo/Robot_logo.pdf} Robot,
\includegraphics[height=0.9em]{logo/Simulation_logo.pdf} Simulation,

\textbf{Sensor:} 
\badge{orange}{G}GelSight,
\badge{gray}{D}DIGIT,
\badge{teal}{T}Tac3D,
\badge{pink}{G}GelStereo,
\badge{blue}{S}Simulated

\textbf{Category:} T-V: Tactile-Vision;
T-L: Tactile-Language;
T-V-L: Tactile-Vision-Language;
T-V-O: Tactile-Vision-Other.
}
\label{tab:vision_tactile_grasping_datasets}
\end{table*}

\subsection{Tactile-Vision Datasets}
\label{subsec:vt_dataset}

T-V datasets form an important empirical basis for multimodal tactile research by pairing visual observations with contact-induced tactile responses. Early datasets, including the VT dataset~\cite{liu2016visual}, Feeling of Success~\cite{calandra2017feeling}, CLAF~\cite{yuan2017connecting}, and ACVTM~\cite{calandra2018more}, mainly focused on controlled robotic grasping with high-resolution sensors such as GelSight to reduce visual ambiguity in object recognition and grasp prediction. To support cross-modal synthesis and joint representation learning, later datasets such as Multimodal Grasp~\cite{wang2019multimodal} and VisGel~\cite{li2019connecting} expanded the diversity of interactions across household objects. More recently, the paradigm has shifted toward less constrained and more complex physical interactions. Touch and Go~\cite{yang2022touch} captures free-form human exploration in the wild, whereas TouchClothing~\cite{gao2023controllable} and CFV~\cite{li2023learning} target specialized interactions with deformable garments and articulated objects, respectively. This progression from controlled laboratory settings to more realistic environments is further reflected in recent large-scale benchmarks such as Touch in the Wild~\cite{zhu2025touch} and TacQuad~\cite{feng2025anytouch}, which provide large amounts of aligned multi-sensor data for generalized T-V fusion.

\subsection{Tactile-Language Datasets}
\label{subsec:lt_dataset}

T-L datasets directly align tactile observations with natural language, enabling semantic grounding without relying on visual observations. Existing datasets mainly map tactile perception to property-level or concept-level semantics. For example, PhysiCLeAR~\cite{yu2024octopi} annotates GelSight videos with human-described physical attributes such as hardness and roughness to support material understanding. Extending to spatial representations, TCL3D~\cite{ma2025cltp} pairs 3D tactile point clouds and rendered images with language descriptions to ground contact geometry explicitly. Moving beyond predefined attributes, STOLA~\cite{cheng2025stola} associates tactile signals with open-ended and commonsense descriptions to support more advanced tactile reasoning.

Compared with vision-centered datasets, T-L datasets place greater emphasis on semantic abstraction, interpretability, and linguistic supervision. Although they remain limited in scale, they are important for tasks such as tactile captioning and language-grounded representation learning, bridging the critical gap between physical transduction and high-level semantic reasoning.

\subsection{Tactile-Vision-Language Datasets}
\label{subsec:vlt_dataset}

T-V-L datasets combine visual observations, tactile signals, and natural language annotations, providing the most semantically expressive setting for multimodal tactile fusion. By jointly aligning perception, contact, and language, they support cross-modal alignment, multimodal reasoning, and language-grounded tactile representation learning.

Early T-V-L datasets mainly establish tri-modal alignment in controlled settings. The TVL dataset~\cite{fu2024touch} and the TLV dataset~\cite{cheng2024towards} align DIGIT-based tactile signals, RGB images, and sentence-level descriptions through structured annotation pipelines. MMWand~\cite{chi2024multi} extends this setting to multimodal interaction sequences collected with a handheld device equipped with visual, force, and tactile sensors, together with human-annotated language descriptions.

More recent datasets improve scale, semantic richness, and temporal coverage. Touch100k~\cite{cheng2025touch100k} expands T-V-L alignment to more than 100{,}000 samples and provides both fine-grained phrases and longer natural language descriptions. The dataset introduced in DAMF~\cite{wang2025damf} provides high-quality tri-modal data under controlled interactions for evaluating T-V-L fusion. VTV150K~\cite{xie2025universal} introduces large-scale visual-tactile video data with tactile attribute annotations, enabling temporal modeling, while MSDO~\cite{li2025tvt} provides standardized T-V-L triplets for object-level recognition under real-world variability. In addition, ImageNet-T~\cite{cho2025ra}, although it does not include physical tactile measurements, serves as a tactile-aware vision-language resource by enriching images with material- and texture-oriented descriptions.

Overall, T-V-L datasets show a clear shift toward larger, semantically richer, and more temporally structured datasets that better connect perception, language, and interaction.

\subsection{Tactile-Vision-Other Datasets}
\label{subsec:vto_dataset}

T-V-O datasets extend T-V settings by incorporating additional modalities such as audio, action, force, or proprioception. Here, ``Other'' denotes signals that capture interaction dynamics, embodiment, or environmental feedback beyond static T-V perception.

An early representative line is the ObjectFolder series~\cite{gao2021objectfolder,gao2022objectfolder,gao2023objectfolder}, which augments vision and tactile sensing with impact audio for multisensory object understanding. The earlier versions focus on large-scale simulated objects, while ObjectFolder-Real provides densely aligned real-world measurements. These datasets mainly strengthen object-centric perception by introducing complementary sensory channels for cross-modal learning and sim-to-real transfer.

Later T-V-O datasets place greater emphasis on embodied interaction. TLA~\cite{hao2025tla} combines sequential tactile observations with language-conditioned manipulation trajectories, supporting the study of contact-rich control and action-conditioned tactile perception. VinT-6D~\cite{wan2024vint} further incorporates proprioception together with vision and touch, enabling research on dexterous manipulation and occlusion-robust object-in-hand perception. OmniViTac~\cite{zheng2026omnivtavisuotactileworldmodeling} extends this trend by providing a large-scale visuo-tactile-action dataset for contact-rich manipulation, with diverse tasks, objects, and temporally aligned multimodal streams. Other recent datasets expand the interaction setting from localized contact to richer manipulation processes. OPENTOUCH~\cite{song2025opentouch} focuses on full-hand tactile sensing during natural object manipulation, FreeTacMan~\cite{wu2025freetacman} supports scalable T-V data collection in diverse contact-rich scenarios, and Hoi!~\cite{engelbracht2025hoi} integrates force-grounded articulated manipulation with cross-view visual observation.

Overall, T-V-O datasets broaden multimodal tactile fusion from multisensory perception to embodied perception-action modeling by introducing dynamic and physically grounded signals beyond vision and touch.
\section{Multimodal Methods}
\label{sec:methods}

\begin{figure}[]
    \centering
    \includegraphics[width=0.5\textwidth]{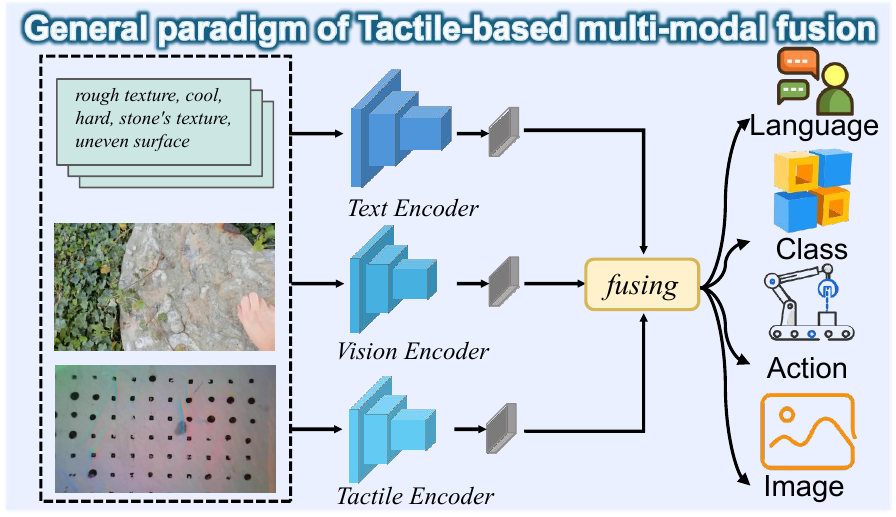}
    \vspace{-4mm}
    \caption{General paradigm of multimodal tactile fusion with downstream tasks.}
    \vspace{-4mm}
    \label{fusing}
\end{figure}

Multimodal methods form the algorithmic core of multimodal tactile fusion, determining how heterogeneous sensory signals are encoded, aligned, fused, and used. As illustrated by the unified encoder-fusion-task pipeline in Fig.~\ref{fusing}, modality-specific inputs are mapped into learned representations and integrated to support downstream tasks. We categorize the existing literature into three primary paradigms: \textbf{(1) Multimodal Perception and Recognition}; \textbf{(2) Multimodal Cross-Modal Generation and Transformation}; and \textbf{(3) Multimodal Interaction and Manipulation}. Table~\ref{tab:method_summary} summarizes representative methods in these paradigms. In the following subsections, we review these paradigms with an emphasis on their fusion strategies and learning objectives.

\begin{table*}[]
\centering
\caption{Summary of multimodal tactile fusion methods published before the first quarter of 2026. }

\renewcommand{\arraystretch}{1.15}
\setlength{\tabcolsep}{4pt}
\resizebox{1.0\textwidth}{!}{
\begin{tabular}{l c l c c c| c c c c | c c | c c c}
\hline 
\multirow{2}{*}{\textbf{Method}} &
\multirow{2}{*}{\textbf{Year}} &
\multirow{2}{*}{\textbf{Input Modality}} &
\multirow{2}{*}{\includegraphics[height=0.9em]{logo/vision_logo.pdf}} &
\multirow{2}{*}{\includegraphics[height=0.9em]{logo/touch_logo.pdf}} &
\multirow{2}{*}{\includegraphics[height=0.9em]{logo/language_logo.pdf}} &
\multicolumn{4}{c|}{\textbf{PR}} &
\multicolumn{2}{c|}{\textbf{CG}} &
\multicolumn{2}{c}{\textbf{IM}} \\
\cline{7-14}
& & & & & & 
\textbf{OR} & \textbf{AM} & \textbf{GS} & \textbf{RM} &
\textbf{VTG} & \textbf{TLG} &
\textbf{RP}  & \textbf{LM} \\
\hline

VT~\cite{liu2016visual}
& 2016 & \includegraphics[height=0.9em]{logo/vision_logo.pdf} \includegraphics[height=0.9em]{logo/touch_logo.pdf} & 
- & -  & -
& \checkmark &  
&  &  &  &  &  & \\

CLAF~\cite{yuan2017connecting}
& 2017 & \includegraphics[height=0.9em]{logo/vision_logo.pdf} \includegraphics[height=0.9em]{logo/touch_logo.pdf} & 
CNN~\cite{krizhevsky2012imagenet} & CNN~\cite{krizhevsky2012imagenet} & -
&  & 
&  & \checkmark &  &  &  & \\

TFOS~\cite{calandra2017feeling}
& 2017 & \includegraphics[height=0.9em]{logo/vision_logo.pdf} \includegraphics[height=0.9em]{logo/touch_logo.pdf} &
ImageNet~\cite{krizhevsky2012imagenet} & ImageNet~\cite{krizhevsky2012imagenet} & -
&  &  & \checkmark & 
&  & 
&  &    \\

ACVTM~\cite{calandra2018more}
& 2018 & \includegraphics[height=0.9em]{logo/vision_logo.pdf} \includegraphics[height=0.9em]{logo/touch_logo.pdf}
\includegraphics[height=0.9em]{logo/action_logo.pdf}& 
ResNet~\cite{he2016deep} & ResNet~\cite{he2016deep} & -
&  & 
& \checkmark  
&  &  & 
& \checkmark 
& \\

ViTac~\cite{luo2018vitac}
& 2018 & \includegraphics[height=0.9em]{logo/vision_logo.pdf} \includegraphics[height=0.9em]{logo/touch_logo.pdf} & 
AlexNet~\cite{krizhevsky2012imagenet} & CNN~\cite{krizhevsky2012imagenet} & -
&  
& \checkmark 
&  &  &  &  &  & \\

SFA~\cite{fazeli2019see}
& 2019 & \includegraphics[height=0.9em]{logo/vision_logo.pdf} \includegraphics[height=0.9em]{logo/touch_logo.pdf} & 
Mask R-CNN~\cite{he2017mask} & BNN~\cite{kononenko1989bayesian} & -
&  &  &  &  &  & 
& \checkmark
& \\

LVTR~\cite{li2019learning}
& 2019 & \includegraphics[height=0.9em]{logo/vision_logo.pdf} \includegraphics[height=0.9em]{logo/touch_logo.pdf} & 
ResNet~\cite{he2016deep} & GAN~\cite{goodfellow2014generative} & -
&  &  &  &
& \checkmark & 
& & \\

Making sense~\cite{lee2019making}
& 2019 & \includegraphics[height=0.9em]{logo/vision_logo.pdf} \includegraphics[height=0.9em]{logo/touch_logo.pdf} 
\includegraphics[height=0.9em]{logo/action_logo.pdf} & 
FlowNet~\cite{dosovitskiy2015flownet} & CNN~\cite{krizhevsky2012imagenet} & -
&  &  &  &  &  & 
& \checkmark
& \\

CTAV~\cite{li2019connecting}
& 2019 & \includegraphics[height=0.9em]{logo/vision_logo.pdf} \includegraphics[height=0.9em]{logo/touch_logo.pdf} & ResNet~\cite{he2016deep} & ResNet~\cite{he2016deep} & -
&  &  &  & 
& \checkmark & \checkmark
&  &    \\

LLCL~\cite{zheng2020lifelong}
& 2020 & \includegraphics[height=0.9em]{logo/vision_logo.pdf} \includegraphics[height=0.9em]{logo/touch_logo.pdf} & 
VGG-16~\cite{simonyan2014very} & VGG-16~\cite{simonyan2014very} & -
&  & \checkmark
&  & 
& 
&  &  & \\


VTT~\cite{chen2022visuo}
& 2022 & \includegraphics[height=0.9em]{logo/vision_logo.pdf} \includegraphics[height=0.9em]{logo/touch_logo.pdf} & ViT~\cite{dosovitskiy2020image} & Transformer~\cite{vaswani2017attention} & -
&  &  &  & 
&  & 
& \checkmark &    \\


TouchAndGo~\cite{yang2022touch}
& 2022 & \includegraphics[height=0.9em]{logo/vision_logo.pdf} \includegraphics[height=0.9em]{logo/touch_logo.pdf} & 
ResNet~\cite{he2016deep} & ResNet~\cite{he2016deep} & -
&  
& \checkmark 
& \checkmark
&  & \checkmark & 
& 
& \\


Visuotactile-RL~\cite{hansen2022visuotactile}
& 2022 & \includegraphics[height=0.9em]{logo/vision_logo.pdf} \includegraphics[height=0.9em]{logo/touch_logo.pdf}
\includegraphics[height=0.9em]{logo/action_logo.pdf}& 
DrQv2~\cite{yarats2021mastering} & DrQv2~\cite{yarats2021mastering} & -
&  &  &  &  &  & 
& \checkmark 
& \\

VITO-Transformer~\cite{li2023vito}
& 2023 & \includegraphics[height=0.9em]{logo/vision_logo.pdf} \includegraphics[height=0.9em]{logo/touch_logo.pdf} & Transformer~\cite{vaswani2017attention} & Transformer~\cite{vaswani2017attention} & -
& \checkmark  &  &  & 
& & 
&  &    \\

UVTS~\cite{li2023learning}
& 2024 & \includegraphics[height=0.9em]{logo/vision_logo.pdf} \includegraphics[height=0.9em]{logo/touch_logo.pdf} & MLP~\cite{tolstikhin2021mlp} & MLP~\cite{tolstikhin2021mlp} & -
&  &  &  
&  & \checkmark & \checkmark
&  &    \\

MViTac~\cite{dave2024multimodal}
& 2024 & \includegraphics[height=0.9em]{logo/vision_logo.pdf} \includegraphics[height=0.9em]{logo/touch_logo.pdf} & ResNet~\cite{he2016deep} & ResNet~\cite{he2016deep} & -
&  & \checkmark & & 
&  & 
&  &    \\

UniTouch~\cite{yang2024binding}
& 2024 & \includegraphics[height=0.9em]{logo/vision_logo.pdf} \includegraphics[height=0.9em]{logo/touch_logo.pdf} \includegraphics[height=0.9em]{logo/language_logo.pdf} & ViT~\cite{dosovitskiy2020image} & ViT~\cite{dosovitskiy2020image} & OpenCLIP~\cite{radford2021learning}
&  & \checkmark & \checkmark & \checkmark
& \checkmark & \checkmark
&  &    \\

Tactile-vlm~\cite{fu2024touch}
& 2024 & \includegraphics[height=0.9em]{logo/vision_logo.pdf} \includegraphics[height=0.9em]{logo/touch_logo.pdf} \includegraphics[height=0.9em]{logo/language_logo.pdf} & ViT~\cite{dosovitskiy2020image} & ViT~\cite{dosovitskiy2020image} & OpenCLIP~\cite{radford2021learning}
&  & \checkmark &  & 
&  & \checkmark
&  &    \\

STLV-Align~\cite{cheng2024towards}
& 2024 & \includegraphics[height=0.9em]{logo/vision_logo.pdf} \includegraphics[height=0.9em]{logo/touch_logo.pdf} \includegraphics[height=0.9em]{logo/language_logo.pdf} & ViT~\cite{dosovitskiy2020image} & ViT~\cite{dosovitskiy2020image} & OpenCLIP~\cite{radford2021learning}
&  & \checkmark &  & 
&  & 
&  &    \\

Octopi~\cite{yu2024octopi}
& 2024 & \includegraphics[height=0.9em]{logo/language_logo.pdf} \includegraphics[height=0.9em]{logo/touch_logo.pdf}  & 
CLIP-ViT~\cite{radford2021learning} & ViFi-CLIP~\cite{rasheed2023fine} & Vicuna1.5~\cite{vicuna2023}
&  & \checkmark &  & 
&  & 
&  &    \\

BVT~\cite{fang2024bidirectional}
& 2024 & \includegraphics[height=0.9em]{logo/vision_logo.pdf} \includegraphics[height=0.9em]{logo/touch_logo.pdf} & 
CLIP~\cite{radford2021learning} & CLIP~\cite{radford2021learning} & -
&  &  &  & 
& \checkmark & 
&  &    \\


Octopi1.5~\cite{yu2025demonstrating}
& 2025 & \includegraphics[height=0.9em]{logo/vision_logo.pdf} \includegraphics[height=0.9em]{logo/language_logo.pdf} \includegraphics[height=0.9em]{logo/touch_logo.pdf}  & Qwen2-VL~\cite{wang2024qwen2} & CLIP-ViT~\cite{radford2021learning} & Qwen2-VL~\cite{wang2024qwen2} 
&  &  &  & 
& \checkmark & 
&  &    \\

ViTacFormer~\cite{heng2025vitacformer}
& 2025 & \includegraphics[height=0.9em]{logo/vision_logo.pdf} \includegraphics[height=0.9em]{logo/touch_logo.pdf}  & Transformer~\cite{vaswani2017attention} & Transformer~\cite{vaswani2017attention} & -
&  & &  & 
&  & 
& \checkmark &    \\

ViTaPEs~\cite{lygerakis2025vitapes}
& 2025 & \includegraphics[height=0.9em]{logo/vision_logo.pdf} \includegraphics[height=0.9em]{logo/touch_logo.pdf}  & Transformer~\cite{vaswani2017attention} & Transformer~\cite{vaswani2017attention} & -
&  & \checkmark & \checkmark & 
&  & 
&  &    \\

STOLA~\cite{cheng2025stola}
& 2025 & \includegraphics[height=0.9em]{logo/language_logo.pdf} \includegraphics[height=0.9em]{logo/touch_logo.pdf}  & 
- & TLV-Link~\cite{cheng2025touch100k} & Vicuna 1.5~\cite{vicuna2023}
&  &  &  & 
&  & \checkmark
&  &    \\

ObjectAttrRecogn~\cite{chen2025object}
& 2025 & \includegraphics[height=0.9em]{logo/vision_logo.pdf} \includegraphics[height=0.9em]{logo/touch_logo.pdf} & ViT~\cite{dosovitskiy2020image} & ViT~\cite{dosovitskiy2020image} & -
&  & \checkmark &  & 
&  & 
&  &    \\

Surformer v1~\cite{kansana2025surformerv1}
& 2025 & \includegraphics[height=0.9em]{logo/vision_logo.pdf} \includegraphics[height=0.9em]{logo/touch_logo.pdf} & ResNet~\cite{he2016deep} & ResNet~\cite{he2016deep} & -
&  & \checkmark &  & 
&  & 
&  &    \\

Surformer v2~\cite{kansana2025surformerv2}
& 2025 & \includegraphics[height=0.9em]{logo/vision_logo.pdf} \includegraphics[height=0.9em]{logo/touch_logo.pdf} & EfficientNetV2~\cite{tan2021efficientnetv2} & Transformer~\cite{vaswani2017attention} & 
&  & \checkmark &  & 
&  & 
&  &    \\

CLTP~\cite{ma2025cltp}
& 2025 & \includegraphics[height=0.9em]{logo/language_logo.pdf} \includegraphics[height=0.9em]{logo/touch_logo.pdf}  & ULIP-2~\cite{xue2024ulip} & CLIP~\cite{radford2021learning} & CLIP~\cite{radford2021learning}
&  & \checkmark  &  & 
& & \checkmark 
&  &    \\

ConViTac~\cite{wu2025convitac}
& 2025 & \includegraphics[height=0.9em]{logo/vision_logo.pdf} \includegraphics[height=0.9em]{logo/touch_logo.pdf} & ViT~\cite{dosovitskiy2020image} & ViT~\cite{dosovitskiy2020image} & 
&  & \checkmark & \checkmark & 
&  & 
&  &    \\

RA-Touch~\cite{cho2025ra}
& 2025 & \includegraphics[height=0.9em]{logo/vision_logo.pdf} \includegraphics[height=0.9em]{logo/touch_logo.pdf} \includegraphics[height=0.9em]{logo/language_logo.pdf} &  & OpenCLIP~\cite{radford2021learning} & OpenCLIP~\cite{radford2021learning}
&  &  &  & 
&  & \checkmark
&  &    \\

TLV-Link~\cite{cheng2025touch100k}
& 2025 & \includegraphics[height=0.9em]{logo/vision_logo.pdf} \includegraphics[height=0.9em]{logo/touch_logo.pdf} \includegraphics[height=0.9em]{logo/language_logo.pdf} & ViT~\cite{dosovitskiy2020image} & ViT~\cite{dosovitskiy2020image} & OpenCLIP~\cite{radford2021learning}
&  & \checkmark & \checkmark & 
&  & 
&  &    \\

TLA~\cite{hao2025tla}
& 2025 & \includegraphics[height=0.9em]{logo/touch_logo.pdf} \includegraphics[height=0.9em]{logo/language_logo.pdf} & - & ViT~\cite{dosovitskiy2020image} & Qwen2~\cite{wang2024qwen2}
&  &  &  & 
&  & 
& \checkmark &    \\

VTLA~\cite{zhang2025vtla}
& 2025 & \includegraphics[height=0.9em]{logo/vision_logo.pdf} \includegraphics[height=0.9em]{logo/touch_logo.pdf} \includegraphics[height=0.9em]{logo/language_logo.pdf} \includegraphics[height=0.9em]{logo/action_logo.pdf} & Qwen-VL~\cite{wang2024qwen2} & ViT~\cite{dosovitskiy2020image} & -
&  &  &  & 
&  & 
& \checkmark &   \\

OmniVTLA~\cite{cheng2025omnivtla}
& 2025 & \includegraphics[height=0.9em]{logo/vision_logo.pdf} \includegraphics[height=0.9em]{logo/touch_logo.pdf} \includegraphics[height=0.9em]{logo/language_logo.pdf} \includegraphics[height=0.9em]{logo/action_logo.pdf} & - & ViT~\cite{dosovitskiy2020image} & ViT~\cite{dosovitskiy2020image}
&  &  &  & 
&  & 
& \checkmark &   \\

VTLG~\cite{li2026vtlg}
& 2025 & \includegraphics[height=0.9em]{logo/vision_logo.pdf} \includegraphics[height=0.9em]{logo/touch_logo.pdf} \includegraphics[height=0.9em]{logo/language_logo.pdf} \includegraphics[height=0.9em]{logo/action_logo.pdf} & ViT~\cite{dosovitskiy2020image} & ViT~\cite{dosovitskiy2020image} & Albert-base-v2
&  &  &  & 
&  & 
&  &  \checkmark \\

VTV-LLM~\cite{xie2025universal}
& 2025 & \includegraphics[height=0.9em]{logo/vision_logo.pdf} \includegraphics[height=0.9em]{logo/touch_logo.pdf} & ViT~\cite{dosovitskiy2020image} & ViT~\cite{dosovitskiy2020image} & -
&  &  &  & 
&  & \checkmark
&  &    \\

TVT-Transformer~\cite{li2025tvt}
& 2025 & \includegraphics[height=0.9em]{logo/vision_logo.pdf} \includegraphics[height=0.9em]{logo/touch_logo.pdf} \includegraphics[height=0.9em]{logo/language_logo.pdf} & Transformer~\cite{vaswani2017attention} & Transformer~\cite{vaswani2017attention} & BERT~\cite{devlin2019bert}
& \checkmark &  &  & 
&  & 
&  &    \\

DT-Transformer~\cite{qiu2024dt}
& 2025 & \includegraphics[height=0.9em]{logo/language_logo.pdf} \includegraphics[height=0.9em]{logo/touch_logo.pdf}  & - & CNN~\cite{krizhevsky2012imagenet} & BERT~\cite{devlin2019bert}
& \checkmark &  &  & 
&  & 
&  &    \\

VHTformer~\cite{li2025vhtformer}
& 2025 & \includegraphics[height=0.9em]{logo/vision_logo.pdf} \includegraphics[height=0.9em]{logo/touch_logo.pdf} \includegraphics[height=0.9em]{logo/language_logo.pdf} & CNN~\cite{krizhevsky2012imagenet} & CNN~\cite{krizhevsky2012imagenet} & BERT~\cite{devlin2019bert}
& \checkmark &  &  & 
&  & 
&  &    \\

Touch in the wild~\cite{zhu2025touch}
& 2025 & \includegraphics[height=0.9em]{logo/vision_logo.pdf} \includegraphics[height=0.9em]{logo/touch_logo.pdf} \includegraphics[height=0.9em]{logo/language_logo.pdf} & ViT~\cite{dosovitskiy2020image} & CNN~\cite{krizhevsky2012imagenet} & 
&  &  &  & 
&  & 
& \checkmark &    \\

DAMF~\cite{wang2025damf}
& 2025 & \includegraphics[height=0.9em]{logo/vision_logo.pdf} \includegraphics[height=0.9em]{logo/touch_logo.pdf} \includegraphics[height=0.9em]{logo/language_logo.pdf} & ViT~\cite{dosovitskiy2020image} & DenseNet~\cite{huang2017densely} & -BERT~\cite{devlin2019bert}
& \checkmark &  &  & 
&  & 
&  &    \\

AnyTouch~\cite{feng2025anytouch}
& 2025 & \includegraphics[height=0.9em]{logo/vision_logo.pdf} \includegraphics[height=0.9em]{logo/touch_logo.pdf} & - & - & -
&  & \checkmark & \checkmark & 
&  & 
& \checkmark &    \\

TEVG~\cite{tong2025can}
& 2025 & \includegraphics[height=0.9em]{logo/vision_logo.pdf} \includegraphics[height=0.9em]{logo/touch_logo.pdf} & - & - & -
&  &  &  & 
&  & 
& \checkmark &    \\

DreamTacVLA~\cite{ye2025learning}
& 2025 & \includegraphics[height=0.9em]{logo/vision_logo.pdf} \includegraphics[height=0.9em]{logo/touch_logo.pdf} \includegraphics[height=0.9em]{logo/language_logo.pdf} \includegraphics[height=0.9em]{logo/action_logo.pdf} & CLIP-ViT~\cite{radford2021learning} & CLIP-ViT~\cite{radford2021learning} & CLIP-ViT~\cite{radford2021learning}
&  &  &  & 
&  & 
& \checkmark &    \\

VLA-Touch~\cite{bi2025vla}
& 2025 & \includegraphics[height=0.9em]{logo/vision_logo.pdf} \includegraphics[height=0.9em]{logo/touch_logo.pdf} \includegraphics[height=0.9em]{logo/language_logo.pdf} \includegraphics[height=0.9em]{logo/action_logo.pdf} & GPT-4o~\cite{hurst2024gpt} & Octopi~\cite{yu2024octopi} &  GPT-4o~\cite{hurst2024gpt}
&  & \checkmark  &  & 
&  & 
& \checkmark &    \\


DexTac~\cite{zhang2026dextac}
& 2026 & \includegraphics[height=0.9em]{logo/vision_logo.pdf} \includegraphics[height=0.9em]{logo/touch_logo.pdf}  \includegraphics[height=0.9em]{logo/action_logo.pdf} & ResNet~\cite{he2016deep}  & ResNet~\cite{he2016deep}  &  -
&  &  &  & 
&  & 
& \checkmark &    \\


OmniVTA~\cite{zheng2026omnivtavisuotactileworldmodeling}
& 2026 & \includegraphics[height=0.9em]{logo/vision_logo.pdf} \includegraphics[height=0.9em]{logo/touch_logo.pdf}   \includegraphics[height=0.9em]{logo/action_logo.pdf} & ResNet~\cite{he2016deep} & - & -
&  &  &  & 
&  & 
& \checkmark &    \\

VTAM~\cite{yuan2026vtamvideotactileactionmodelscomplex}
& 2026 & \includegraphics[height=0.9em]{logo/vision_logo.pdf} \includegraphics[height=0.9em]{logo/touch_logo.pdf}   \includegraphics[height=0.9em]{logo/action_logo.pdf} & - & - & -
&  &  &  & 
&  & 
& \checkmark &    \\

\hline
\end{tabular}}

\caption*{
\textbf{Input:} 
\includegraphics[height=0.9em]{logo/touch_logo.pdf} touch,
\includegraphics[height=0.9em]{logo/action_logo.pdf} action,
\includegraphics[height=0.9em]{logo/language_logo.pdf} language,
\includegraphics[height=0.9em]{logo/vision_logo.pdf} vision. \\
\textbf{Action:} PR: Multimodal Perception and Recognition;
CG: Cross-Modal Generation and Transformation;
IM: Multimodal Interaction and Manipulation;
OR: Object Recognition;
AM: Attribute and Material Recognition;
GS: Grasp Success/Failure Prediction;
RM: Cross-Modal Retrieval and Matching;
VTG: Vision-Tactile Generation;
TLG: Tactile-Language Generation;
RP: Robot Manipulation with Multimodal Perception;
LM: Language-Guided Manipulation. \\

}

\label{tab:method_summary}
\end{table*}

\subsection{Multimodal Perception and Recognition}

\begin{figure}[]
    \centering
    \includegraphics[width=0.5\textwidth]{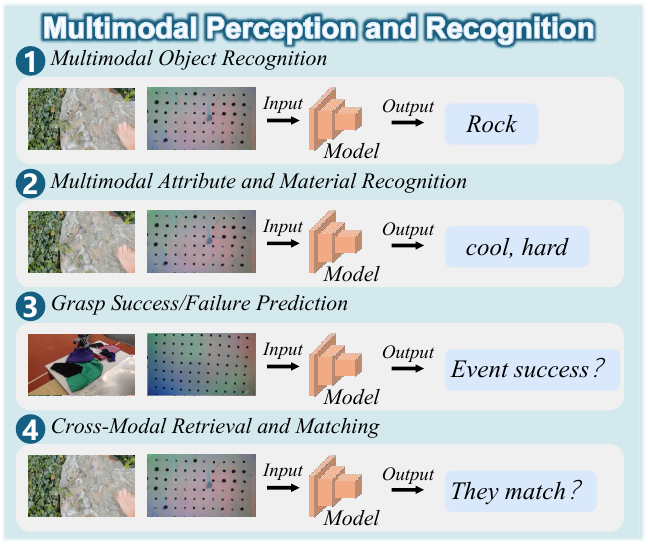}
    \caption{Categorization of multimodal perception and recognition, including multimodal object recognition (Section~\ref{sec:MOR}), multimodal attribute and material recognition (Section~\ref{sec:MAMR}), grasp success or failure prediction (Section~\ref{sec:GSFP}), and cross-modal retrieval and matching (Section~\ref{sec:CMRM}).}
    \label{sec4.1}
\end{figure}

Multimodal perception and recognition is the most widely studied paradigm in multimodal tactile fusion. Its primary goal is to infer semantic and functional properties by integrating tactile signals with complementary modalities, mainly vision and language. As shown in Fig.~\ref{sec4.1}, these methods typically follow an encode-fuse-predict pipeline that maps heterogeneous observations to task-specific outputs.

We categorize this paradigm into four tasks: object recognition, attribute and material recognition, grasp success or failure prediction, and cross-modal retrieval and matching. Although these tasks often share similar fusion backbones, they differ substantially in supervision signals, evaluation protocols, and the specific role of tactile feedback. The following subsections review representative methods for these tasks, with an emphasis on their architectural designs and fusion strategies.

\subsubsection{Multimodal Object Recognition}
\label{sec:MOR}

Multimodal object recognition identifies object categories by integrating complementary cues: vision for global appearance, tactile sensing for local physical properties, and language for high-level semantic priors. The main challenge is to align these heterogeneous modalities, which differ in spatial structure, noise characteristics, and semantic granularity, while preserving their complementary strengths.

\textbf{Early T-V Fusion Methods:} Early research mainly focused on T-V fusion. A representative framework was introduced by Liu \emph{et al.}~\cite{liu2016visual}, showing that augmenting visual features with tactile sequences improves robustness under visual ambiguity. To address weakly paired settings without strict sample-level correspondence, a joint group kernel sparse coding method with an $\ell_{2,1}$-norm constraint was proposed by Liu and Sun~\cite{liu2018visual} to enforce class-level sparsity consistency. These studies demonstrated the practical value of T-V Fusion.

\textbf{Deep Cross-Modal Architectures:} Subsequent methods moved toward deep cross-modal architectures. VITO-Transformer~\cite{li2023vito} introduced attention-based fusion to jointly model global vision features and local tactile features, and later methods further incorporated language inputs. For example, DT-Transformer~\cite{qiu2024dt} introduced text as an auxiliary semantic modality to reduce ambiguity in sparse tactile signals, while TVT-Transformer~\cite{li2025tvt} extended this direction to a unified T-V-L architecture.

\textbf{Recent T-V-L Fusion and Spatial Alignment:} Recent studies place greater emphasis on flexible T-V-L fusion and fine-grained spatial alignment. VHTformer~\cite{li2025vhtformer} employs a joint-query mechanism to support symmetric vision-tactile-language interaction and mitigate modality bias. DAMF~\cite{wang2025damf} uses language-guided attention to improve T-V fusion, especially when tactile data is limited. In addition, ViTaPEs~\cite{lygerakis2025vitapes} improves cross-modal alignment through position encodings that explicitly model the spatial relationship between vision observations and physical contacts. Overall, this line of research shows a clear transition from basic T-V pairing to tighter T-V-L integration, where language increasingly serves as an important semantic scaffold.

\subsubsection{Multimodal Attribute and Material Recognition}
\label{sec:MAMR}

Multimodal attribute and material recognition infers fine-grained physical properties, such as texture, compliance, and contact geometry, by integrating tactile signals with vision and language cues. Unlike global object recognition, this task relies more heavily on local and contact-aware semantics, where vision alone is often insufficient and tactile feedback provides important complementary information.

\textbf{Early Shared T-V Representations:} Early methods mainly focused on learning shared representations across modalities. Representative frameworks in ViTac~\cite{luo2018vitac} and a joint sparse coding approach were introduced to demonstrate that cross-modal fusion improves material identification in ambiguous or weakly paired settings, with the latter proposed by Liu and Sun~\cite{liu2018visual}. This line of research was later extended to continual learning to reduce catastrophic forgetting by LLCL~\cite{zheng2020lifelong} and to larger-scale self-supervised cross-modal association with human-collected datasets by Touch and Go~\cite{yang2022touch}.

\textbf{Transferable Learning and Semantic Grounding:} More recent studies have shifted toward scalable representation learning enabled by foundation models. Methods such as UniTouch~\cite{yang2024binding} and MViTac~\cite{dave2024multimodal} use contrastive learning to align tactile features with pre-trained vision models, enabling zero-shot material inference without explicit task-specific supervision. Meanwhile, language has become an important semantic bridge. Frameworks such as Octopi~\cite{yu2024octopi}, Tactile-VLM~\cite{fu2024touch}, STLV-Align~\cite{cheng2024towards}, and CLTP~\cite{ma2025cltp} ground tactile observations in language descriptions, enabling open-vocabulary attribute reasoning. To further improve this alignment, Transformer-based architectures, including Surformer~\cite{kansana2025surformerv1,kansana2025surformerv2}, ConViTac~\cite{wu2025convitac}, and ViTaPEs~\cite{lygerakis2025vitapes}, explicitly model the spatial and semantic consistency between global vision observations and localized physical contacts.

\textbf{Sensor-Agnostic and Scalable Architectures:} More recently, this line of research has increasingly addressed data scarcity and hardware heterogeneity. Representative advances include retrieval-augmented learning that transfers vision material priors to tactile perception (RA-Touch~\cite{cho2025ra}), multitask joint prediction~\cite{chen2025object}, and unified static-dynamic representations that generalize across heterogeneous tactile sensors (AnyTouch~\cite{feng2025anytouch}). Large-scale frameworks such as TLV-Link~\cite{cheng2025touch100k} further illustrate this trend. Overall, this area has progressed from task-specific T-V feature sharing to more unified and sensor-agnostic frameworks with stronger language grounding and cross-modal alignment.

\subsubsection{Grasp Success/Failure Prediction}
\label{sec:GSFP}

Grasp success or failure prediction evaluates grasp stability after contact and during lifting, relying primarily on real-time sensory feedback rather than pre-grasp planning. Because this task depends strongly on contact-rich cues, such as local geometry, pressure distribution, and incipient slip, multimodal approaches integrate vision and tactile signals to overcome the occlusions and limited observability associated with single-modality perception.

\textbf{Foundational T-V Assessment:} Early studies established the importance of tactile feedback in post-contact grasp evaluation. Representative frameworks such as TFOS~\cite{calandra2017feeling} demonstrated that integrating tactile sequences significantly outperforms vision-only prediction, especially when visual evidence becomes unreliable after contact. Following this line of work, ACVTM~\cite{calandra2018more} highlighted the value of T-V fusion not only for success prediction but also for closed-loop regrasping, emphasizing the need for contact-aware manipulation.

\textbf{Transferable Representations and Scalability:} More recent research has shifted from task-specific predictors toward more general and transferable multimodal representations. Models trained on large-scale cross-modal associations, such as Touch and Go~\cite{yang2022touch}, can transfer effectively to grasp assessment. In addition, foundation-model-based architectures such as MViTac~\cite{dave2024multimodal} and UniTouch~\cite{yang2024binding} show that aligned cross-modal representations can support grasp prediction with limited supervision or even zero-shot generalization.

\textbf{Robustness and Fine-Grained Alignment:} Recent frameworks further improve spatial alignment and robustness across heterogeneous tactile sensors. ConViTac~\cite{wu2025convitac} enhances vision-tactile consistency through contrastive alignment for fine-grained interaction reasoning. Meanwhile, models such as TLV-Link~\cite{cheng2025touch100k}, AnyTouch~\cite{feng2025anytouch}, and ViTaPEs~\cite{lygerakis2025vitapes} improve scalability across different data scales, sensor morphologies, and evaluation settings. Overall, this line of work has advanced grasp success prediction into an important downstream task for evaluating robust multimodal tactile perception.

\subsubsection{Cross-Modal Retrieval and Matching}
\label{sec:CMRM}

Cross-modal retrieval and matching align tactile signals with vision or language through shared representations, enabling the retrieval of semantically related counterparts across modalities. In multimodal tactile fusion, this task mainly serves as a benchmark for evaluating whether learned representations preserve contact-aware semantics while supporting robust cross-modal transfer.

\textbf{Retrieval-Oriented Representation Alignment:} Early frameworks such as CLAF~\cite{yuan2017connecting} learned shared T-V representations to support bidirectional retrieval for material understanding. More recent approaches increasingly treat retrieval not only as a standalone task but also as an evaluation protocol for general-purpose multimodal representations. For example, UniTouch~\cite{yang2024binding} demonstrates that aligned tactile features can support zero-shot tactile-to-image and tactile-to-text retrieval, linking physical contact with broader vision and language semantics.

\textbf{Benchmarking Cross-Modal Transferability:} Although dedicated retrieval-only architectures remain relatively limited, this paradigm plays an important role in measuring the alignment quality, semantic richness, and transferability of modern multimodal tactile models.

\subsection{Cross-Modal Generation and Transformation}
\label{subsec:cross_generation}

\begin{figure}[]
    \centering
    \includegraphics[width=0.5\textwidth]{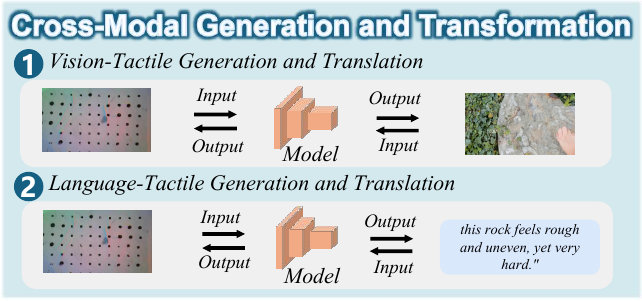}
    \vspace{-4mm}
    \caption{Categorization of multimodal cross-modal generation and transformation, including vision-tactile generation and translation (Section~\ref{sec:VTGT}) and language-tactile generation and translation (Section~\ref{sec:LTGT}).}
    \vspace{-4mm}
    \label{sec4.2}
\end{figure}

Cross-modal generation and transformation expand multimodal tactile fusion from recognition to cross-sensory synthesis. As shown in Fig.~\ref{sec4.2}, these methods mainly follow a conditional generation paradigm, learning mappings between modalities to synthesize target signals. Unlike perception tasks, they place greater emphasis on semantic alignment, structural consistency, and faithful reconstruction.

We divide this paradigm into two directions:
\textbf{(1) Visual-Tactile Generation and Translation:} This line studies direct visual-tactile correspondence, generating local tactile responses from visual inputs, or visual signals from tactile observations.
\textbf{(2) Language-Tactile Generation and Translation:} This line introduces language, enabling tactile signals to be described in natural language or generated from semantic prompts.

Although both directions often rely on encoder-decoder architectures and cross-modal alignment objectives, they differ in output space and semantic granularity. The following subsections review the main methods in these two directions.

\subsubsection{Vision-Tactile Generation and Translation}
\label{sec:VTGT}

Vision-tactile generation and translation investigates bidirectional synthesis between vision observations and tactile responses by modeling their underlying cross-modal correspondences. Unlike discriminative tasks, this generative setting emphasizes cross-modal reconstruction and the preservation of shared physical structures, such as texture, contact geometry, and material properties.

\textbf{Foundational Supervised Synthesis:} Early studies mainly explored supervised and often unidirectional generation. For example, LVTR~\cite{li2019learning} established the feasibility of the feasibility of structured vision-to-tactile translation by converting texture images into tactile vibrations using an ensemble GAN. Extending beyond one-way mapping, CTAV~\cite{li2019connecting} introduced bidirectional prediction by embedding both vision and tactile signals into shared representations, enabling reciprocal cross-modal inference.

\textbf{Scalable and Unified Representations:} Subsequent methods shifted toward more scalable and representation-driven architectures. Models trained on large-scale cross-modal associations, such as Touch and Go~\cite{yang2022touch} can support bidirectional generation alongside recognition. This direction was further advanced by BVT~\cite{fang2024bidirectional}, which constructs explicit bidirectional mappings, and UVTS~\cite{li2023learning}, which preserves critical spatial and contact-aware semantics during translation. More recently, UniTouch~\cite{yang2024binding} has further advanced this trend by aligning tactile features with pre-trained multimodal foundation models, enabling more general vision-tactile generation without heavy reliance on paired task-specific supervision.

Overall, this research direction has progressed from isolated supervised translation pipelines to more unified representation frameworks. Recent approaches increasingly treat high-quality vision-tactile generation not only as a standalone task, but also as a consequence of robust and scalable multimodal alignment.

\subsubsection{Language-Tactile Generation and Translation}
\label{sec:LTGT}

\paragraph{Tactile-to-Language Description Generation}

Tactile-to-language generation translates tactile signals into natural language descriptions, such as texture, compliance, and local geometry, helping bridge the gap between raw physical contact and semantic reasoning. Unlike visual captioning, this task focuses on contact-centered semantics, making tactile perception more interpretable for human-robot communication and high-level decision-making.

\textbf{Foundational Cross-Modal Alignment:} Early frameworks approached tactile description generation mainly through joint representation learning. For example, Tactile-VLM~\cite{fu2024touch} maps sensory inputs into a unified vision-language space, while UniTouch~\cite{yang2024binding} leverages pre-trained vision-language models to support zero-shot description generation without task-specific supervision. These studies show that robust cross-modal alignment is a key prerequisite for tactile-to-language generation.

\textbf{Grounded and Interactive Generation:} More recent methods have shifted toward explicitly grounded and open-ended language generation. CLTP~\cite{ma2025cltp} decodes fine-grained contact states, such as shape and depth, directly from tactile arrays, whereas VTV-LLM~\cite{xie2025universal} extends this capability to temporal vision-tactile streams for dynamic interactions. To improve semantic richness, recent architectures increasingly incorporate retrieval-augmented mechanisms. RA-Touch~\cite{cho2025ra} uses external vision-language knowledge to refine descriptions, and Octopi-1.5~\cite{yu2025demonstrating} adopts lightweight retrieval for real-time and context-aware responses in interactive settings. In addition, SToLa~\cite{cheng2025stola} moves beyond rigid template-based captions toward free-form descriptions and tactile commonsense reasoning.

Overall, this line of work shows a clear transition from basic representation alignment to more interactive and retrieval-augmented generation, where language increasingly serves as a high-level interface for multimodal tactile understanding.

\paragraph{Text-Driven Tactile Synthesis}

Text-driven tactile synthesis maps language directly to physically grounded tactile signals, allowing systems to infer contact sensations from semantic prompts. The main challenge is to connect abstract language concepts, such as texture, compliance, and contact state, to localized physical representations without relying on vision observations.

\textbf{Emerging Text-to-Tactile Generation:} Dedicated research in this direction is still limited. However, representative frameworks such as UniTouch~\cite{yang2024binding} show the feasibility of this direction by anchoring tactile features in pre-trained vision-language representation spaces. Based on these aligned embeddings, textual prompts can guide existing generative models to achieve zero-shot text-to-tactile synthesis without paired task-specific supervision. These early results suggest that tactile synthesis can emerge naturally from large-scale multimodal alignment, with language serving as a flexible high-level interface for reasoning about diverse contact scenarios.

\subsection{Multimodal Interaction and Manipulation}

\begin{figure}[]
    \centering
    \includegraphics[width=0.5\textwidth]{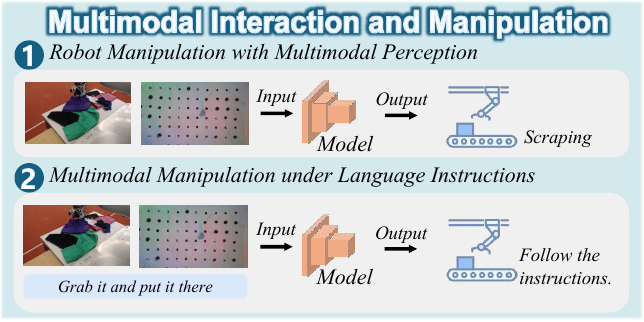}
    \vspace{-4mm}
    \caption{Categorization of multimodal interaction and manipulation, including robot manipulation with multimodal perception (Section~\ref{sec:RMMP}) and multimodal manipulation under language instructions (Section~\ref{sec:MMLI}).}
    \vspace{-4mm}
    \label{sec4.3}
\end{figure}

Multimodal interaction and manipulation is the most application-oriented branch of multimodal tactile fusion, where perception is tightly coupled with physical action. As shown in Fig.~\ref{sec4.3}, these methods extend multimodal fusion from representation learning to closed-loop control, using fused sensory features to guide robot actions in contact-rich and dynamic environments. Compared with recognition or generation tasks, they must also account for action feasibility, contact dynamics, and real-time feedback.

Existing studies can be grouped into two settings: robot manipulation with multimodal perception and multimodal manipulation under language instructions. The former mainly combines tactile and vision signals to improve robustness and precision in tasks such as grasping, insertion, and surface interaction. The latter further introduces language for high-level task guidance, while tactile feedback grounds instruction execution during contact. In the following, we review representative methods in these two settings and discuss how multimodal fusion improves control performance and generalization.

\subsubsection{Robot Manipulation with Multimodal Perception}
\label{sec:RMMP}

\paragraph{Stable Grasping Control with Tactile Feedback}

Stable grasping control uses continuous contact feedback to maintain grasp stability during execution. Unlike pre-grasp planning based mainly on vision cues, it focuses on post-contact adjustments driven by tactile signals, such as force distribution and incipient slip, which are particularly important when handling transparent, reflective, or deformable objects under visual ambiguity.

\textbf{Early Closed-Loop T-V Control:} Early frameworks such as ACVTM~\cite{calandra2018more} demonstrated the value of closed-loop T-V refinement, enabling iterative pose and force adjustment to recover from unstable grasps without applying excessive force. Building on this direction, Visuotactile-RL~\cite{hansen2022visuotactile} formulated stable grasping as a reinforcement learning problem and showed improved robustness and object generalization. Similarly, TEVG~\cite{tong2025can} highlights tactile feedback as an important complement when vision becomes unreliable in transparent object manipulation.

\textbf{Recent Multimodal Policy Learning:} More recent work has extended this setting toward unified multimodal policy learning. OmniVTLA~\cite{cheng2025omnivtla} integrates semantically aligned tactile tokens into an action generation backbone, supporting contact-aware policy learning for complex pick-and-place tasks. Overall, this line of research shows a transition from reactive grasp stabilization to more unified multimodal policy learning for robust manipulation.

\paragraph{Fine Manipulation and Assembly Tasks}
Fine manipulation and assembly tasks require precise perception of continuous contact states, surface interactions, and dynamic forces. In these contact-rich settings, T-V fusion is especially important because visual occlusion and contact uncertainty often make unimodal control unreliable.

\textbf{Early Multimodal Representation Learning:} Early approaches mainly focused on multimodal representation learning. Self-supervised vision-force fusion was used by Lee \emph{et al.}~\cite{lee2019making} to improve sample efficiency in peg-insertion tasks. In addition, a hierarchical vision-tactile framework was introduced by SFA~\cite{fazeli2019see} to infer latent interaction modes, enabling fine-grained control in assembly tasks such as Jenga.

\textbf{Attention-Based Cross-Modal Fusion:} Later methods increasingly adopted explicit cross-modal attention mechanisms. VTT~\cite{chen2022visuo} showed that tactile feedback can guide vision attention toward task-relevant regions during pushing and picking. ViTacFormer~\cite{heng2025vitacformer} extended this Transformer-based design to dexterous manipulation, demonstrating the effectiveness of attention-based modeling in more complex scenarios.

\textbf{Contact-Aware Policies and Broader Integration:} Recent research places greater emphasis on contact-aware policies and broader multimodal integration. DexTac~\cite{zhang2026dextac} uses explicit contact-region and force cues for high-precision tasks such as syringe injection, while DreamTacVLA~\cite{ye2025learning} predicts future tactile states to stabilize sustained interactions. OmniVTA~\cite{zheng2026omnivtavisuotactileworldmodeling} further introduces a visuo-tactile world model that jointly captures interaction dynamics for contact-rich robotic manipulation. VTAM~\cite{yuan2026vtamvideotactileactionmodelscomplex} integrates tactile sensing into a predictive video world model to support force-aware control in complex contact-rich manipulation. Extending toward vision-language-action settings, VLA-Touch~\cite{bi2025vla} and TLA~\cite{hao2025tla} integrate tactile signals with semantic representations for contact-sensitive reasoning. In addition, unified vision-tactile frameworks across heterogeneous sensors (AnyTouch~\cite{feng2025anytouch}) and portable systems~\cite{zhu2025touch} show promising transferability for fine manipulation in unstructured environments.

Overall, this area has progressed from basic representation learning to more integrated vision-language-action frameworks, showing that dexterous manipulation depends strongly on modeling dynamic contact structures together with higher-level semantic context.

\subsubsection{Multimodal Manipulation under Language Instructions}
\label{sec:MMLI}

Multimodal manipulation under language instructions maps high-level semantic commands to contact-rich physical control. Unlike methods driven only by vision or policy learning, this setting requires joint reasoning over task semantics, object geometry, and dynamic physical interactions, making T-V-L models an important bridge between abstract instructions and low-level execution. 

Recent work increasingly moves toward integrated Tactile-Vision-Language-Action frameworks. VTLA~\cite{zhang2025vtla} incorporates tactile feedback into language-conditioned manipulation to adapt insertion behavior under ambiguous vision, while VTLG~\cite{li2026vtlg} combines language instructions with vision and tactile perception for task-oriented grasp generation. Overall, these studies suggest that effective language-guided manipulation depends on the joint modeling of language instructions, multimodal perception, and continuous physical interaction.
\section{Tactile Sensors for Multimodal Fusion}
\label{sec:tactile_sensors}

Unlike distal vision, tactile perception is inherently shaped by the physical interface between the embodied agent and its environment. Therefore, a sensor's transduction principle, material composition, and mechanical morphology directly determine the observability and semantic richness of contact events. In multimodal tactile fusion, tactile sensors provide the hardware foundation for data acquisition, cross-modal alignment, and downstream learning.

We organize existing tactile sensing platforms for multimodal fusion into four categories: wearable tactile systems, handheld and fingertip sensors, robotic skin and multimodal sensor patches, and gripper-mounted and integrated sensors. These categories span major hardware design choices, ranging from human-centered and localized contact sensing to large-area distributed sensing and end-effector integration, thereby offering a hardware-oriented view of the capabilities and limitations of multimodal tactile fusion systems.

\subsection{Wearable Tactile Systems}

Wearable tactile systems provide an important interface for capturing human interaction priors in multimodal learning. An early representative example is the glove-based platform introduced by Fang et al.~\cite{fang2019glove}, which combines distributed pressure signals with external vision features and shows that spatiotemporal vision-tactile integration improves object recognition over single-modality baselines.

More recent systems place greater emphasis on scalable data collection and transferable skill learning. For example, the OSMO glove~\cite{yin2025osmo} uses distributed 3-axis magnetic sensors to capture full-hand contact and motion dynamics, supporting stronger policy learning than vision-only approaches. Complementing this full-hand design, FreeTacMan~\cite{wu2025freetacman} integrates compact vision-tactile sensors on the thumb and index finger together with accurate pose tracking. This targeted configuration supports efficient large-scale data collection and improves contact-aware policy learning for high-precision tasks such as deformable object manipulation.

\subsection{Handheld and Fingertip Sensors}

Handheld and fingertip sensors provide localized and high-resolution contact perception, making them well suited for direct integration with robotic end-effectors. Unlike large-area tactile skins, these systems emphasize compact form factors and tight cross-modal coupling. A prominent direction is the structural design of vision-tactile fingertips to improve spatial alignment and facilitate downstream fusion. For example, LightTact~\cite{lin2025lighttact} adopts an optical design that reduces reliance on deformation reconstruction, whereas TacThru~\cite{li2025simultaneous} uses a transparent structure to simultaneously capture external visual context and internal tactile deformation, which is useful for manipulating soft, transparent, or occluded objects.

At the same time, these compact interfaces are evolving from purely geometric imagers into richer hardware-level multimodal sensing units. Thermochromic materials were incorporated into a vision-based tactile sensor by Sun et al.~\cite{sun2019novel}, enabling joint perception of force, texture, and thermal cues. Similarly, magnetic sensing was integrated within an elastomer by MagicGel~\cite{shan2025magicgel} to improve force estimation and enable proximity perception, showing that auxiliary physical channels can extend the capabilities of conventional optical designs.

In practical applications, these high-fidelity sensors support both task-specific perception and contact-aware control. Vision-tactile feature fusion was used by Zhang et al.~\cite{zhang2025braille} to improve discrimination of fine embossed structures in braille recognition. For adaptive manipulation, a compact fingertip design is used by FVTS~\cite{zhang2024compact} to support real-time contact estimation and closed-loop grasp control. At the system level, tactile sensing is combined with force measurement, pose tracking, and onboard vision by platforms such as TacUMI~\cite{cheng2026tacumi}, UMI-FT~\cite{choi2026wild}, and Touch in the Wild~\cite{zhu2025touch}. Overall, this trend supports more efficient multimodal data collection and more robust policy learning, making fingertip sensors a practical foundation for contact-rich embodied manipulation.

\subsection{Robotic Skin and Multimodal Sensor Patches}

Unlike localized fingertip sensors, robotic skins and multimodal sensor patches emphasize scalability, mechanical compliance, and distributed multiphysics sensing, such as pressure or slip, to support whole-body contact awareness. 

Early designs integrated multiple sensing mechanisms within compliant structures. For example, a thin-film array was proposed by Mao et al.~\cite{mao2024multimodal} to capture force, temperature, and slip for dexterous grasp control. A hydrogel e-skin was developed by Lv et al.~\cite{lv2025biomimetic} to combine triboelectric and ionic capacitive sensing for dynamic and static material recognition.

To address the wiring bottleneck in large-area deployments, piezoresistive sponges were combined with electrical resistance tomography (ERT) by Zheng et al.~\cite{zheng2024large,zheng2025large}. This architecture supports scalable contact localization, force estimation, and touch-type discrimination over large collaborative robot surfaces without dense wiring.

More recent studies increasingly explore adaptive and neuromorphic sensor designs. Self-powered dual-function sensors with in-sensor computation and tactile-pain reflexes were introduced by Wu et al.~\cite{wu2025biomimetic} and Li et al.~\cite{li2025pollen}, supporting autonomous self-protective grasping. Extending this direction, optical, triboelectric, and inertial sensing mechanisms are integrated by SuperTac~\cite{li2026biomimetic} to enrich contact interpretation through multimodal fusion. Overall, these developments extend tactile fusion from localized point contact to more distributed body-level perception, which is important for safe and adaptive human-robot interaction.

\subsection{Gripper-Mounted and Integrated Sensors}

Gripper-mounted systems integrate multimodal perception directly at the manipulation interface, providing compact and co-located feedback for closed-loop control. To maintain consistent observations across demonstration and execution, tactile modules are integrated by modular platforms such as TacUMI~\cite{cheng2026tacumi} and Touch in the Wild~\cite{zhu2025touch} for real-world contact-rich tasks. Another practical direction is to retrofit standard grippers with tactile sensors. Plug-and-play fingertip replacements are provided by LightTact~\cite{lin2025lighttact} and TacThru~\cite{li2025simultaneous}, offering high-resolution and transparent vision-tactile imaging, respectively. To extend sensing channels with limited hardware modification, flexible films were applied to parallel-jaw grippers by Mao et al.~\cite{mao2024multimodal} for slip and thermal perception, while RGB cameras and pressure sensors were embedded into two-finger grippers by Ye et al.~\cite{ye2026visual} to support multitask policy learning. Overall, end-effector-level integration improves contact interpretation, control precision, and the efficiency of multimodal learning.
\section{Evaluation and Comparative Analysis}
\label{sec:evaluation}

Unlike more established multimodal areas such as vision-language learning, multimodal tactile fusion still lacks unified benchmarks and standardized evaluation protocols. Existing evaluation practices remain highly task-dependent, spanning perception, recognition, cross-modal generation, and physical manipulation. This diversity reflects the heterogeneous nature of tactile sensing, variations in sensor hardware, and the close coupling between tactile signals and physical interaction contexts. As a result, most of the studies adopt evaluation protocols tailored to specific tasks and application settings rather than following a common benchmarking pipeline. Understanding these evaluation settings is therefore important for interpreting reported results and enabling more meaningful comparisons across different research directions.

\begin{table*}[!htbp]
\centering
\caption{Summary of evaluation metrics used in multimodal tactile fusion.}
\label{tab:metrics_summary}
\renewcommand{\arraystretch}{1.25}
\setlength{\tabcolsep}{3.5pt}
\resizebox{\textwidth}{!}{
\begin{tabular}{c c p{4.2cm} p{6cm} p{4.6cm} p{3.0cm}}
\hline
\rowcolor{gray!20}
\textbf{Abbr.} & \textbf{$\uparrow/\downarrow$} & \textbf{Full Name} & \textbf{Formula} & \textbf{Description} & \textbf{References} \\
\hline

\rowcolor{blue!10}
\multicolumn{6}{l}{\textbf{Metrics for Multimodal Perception and Recognition}} \\
\hline

ACC & \textcolor{green}{$\uparrow$} & Accuracy 
& $\frac{\mathrm{TP}+\mathrm{TN}}{\mathrm{TP}+\mathrm{TN}+\mathrm{FP}+\mathrm{FN}}$
& Overall proportion of correct predictions.

& \cite{luo2018vitac,li2023vito,qiu2024dt,zheng2020lifelong,calandra2017feeling,wang2025damf,chen2025object,feng2025anytouch} \\

PREC & \textcolor{green}{$\uparrow$} & Precision 
& $\frac{\mathrm{TP}}{\mathrm{TP}+\mathrm{FP}}$
& Correct positives among predicted positives.

& \cite{li2023vito,chen2025object} \\

REC & \textcolor{green}{$\uparrow$} & Recall 
& $\frac{\mathrm{TP}}{\mathrm{TP}+\mathrm{FN}}$
& Correct positives among ground-truth positives.

& \cite{li2023vito,chen2025object} \\

F1 & \textcolor{green}{$\uparrow$} & F1-score 
& $\frac{2 \cdot \mathrm{PREC} \cdot \mathrm{REC}}{\mathrm{PREC}+\mathrm{REC}}$
& Harmonic mean of precision and recall.

& \cite{li2023vito,li2025tvt,wang2025damf,qiu2024dt,chen2025object} \\

mAP & \textcolor{green}{$\uparrow$} & Mean Average Precision 
& $\frac{1}{Q}\sum_{q=1}^{Q}\mathrm{AP}_q$
& Mean AP over queries or classes.

& \cite{li2026vtlg,yang2024binding,cheng2025touch100k,cho2025ra,cheng2024towards} \\

R@k & \textcolor{green}{$\uparrow$} & Recall at $k$ 
& $\frac{1}{Q}\sum_{q=1}^{Q}\mathbb{I}(\text{hit in top-}k)$
& Fraction of queries with a correct match in top-$k$.

& \cite{ma2025cltp,dave2024multimodal,yang2024binding,cheng2025touch100k,cho2025ra,cheng2024towards} \\

MRR & \textcolor{green}{$\uparrow$} & Mean Reciprocal Rank 
& $\frac{1}{Q}\sum_{q=1}^{Q}\frac{1}{\mathrm{rank}_q}$
& Average inverse rank of the first relevant result.

& \cite{cho2025ra} \\

COS & \textcolor{green}{$\uparrow$} & Cosine Similarity 
& $\frac{\mathbf{z}_1 \cdot \mathbf{z}_2}{\|\mathbf{z}_1\|\|\mathbf{z}_2\|}$
& Similarity between two embedding vectors.

& \cite{fu2024touch,yang2024binding,ma2025cltp,wu2025convitac,dave2024multimodal} \\

\hline
\rowcolor{green!10}
\multicolumn{6}{l}{\textbf{Metrics for Cross-Modal Generation}} \\
\hline

MSE & \textcolor{red}{$\downarrow$} & Mean Squared Error 
& $\frac{1}{N}\sum_{i=1}^{N}\|\hat{\mathbf{x}}_i-\mathbf{x}_i\|_2^2$
& Average squared reconstruction error.

& \cite{fang2024bidirectional,li2019connecting} \\

PSNR & \textcolor{green}{$\uparrow$} & Peak Signal-to-Noise Ratio 
& $10\log_{10}\!\left(\frac{L^2}{\mathrm{MSE}}\right)$
& Reconstruction fidelity from MSE.

& \cite{fang2024bidirectional} \\

SSIM & \textcolor{green}{$\uparrow$} & Structural Similarity Index 
& $\frac{(2\mu_x\mu_{\hat{x}}+C_1)(2\sigma_{x\hat{x}}+C_2)}{(\mu_x^2+\mu_{\hat{x}}^2+C_1)(\sigma_x^2+\sigma_{\hat{x}}^2+C_2)}$
& Structural perceptual similarity.

& \cite{fang2024bidirectional} \\

FID & \textcolor{red}{$\downarrow$} & Frechet Inception Distance 
& 
{\parbox[t]{4.2cm}{$\begin{aligned}
\|\mu_r-\mu_g\|_2^2 + {}& \mathrm{Tr}\!\left(\Sigma_r+\Sigma_g\right.\\
&\left.-2(\Sigma_r\Sigma_g)^{1/2}\right)
\end{aligned}$}}

& Distribution gap between real and generated samples.

& \cite{fang2024bidirectional} \\

CA & \textcolor{green}{$\uparrow$} & Classification Accuracy 
& $\frac{N_{\mathrm{correct}}}{N_{\mathrm{all}}}$
& Accuracy of a pre-trained classifier on generated samples.

& \cite{fang2024bidirectional,yang2022touch,fu2024touch,xie2025universal} \\

CVTP & \textcolor{green}{$\uparrow$} & Contrastive visual-Tactile Pretraining Score 
& -
& Embedding alignment between generated outputs and conditions.

& \cite{yang2024binding,ma2025cltp} \\

LLM & \textcolor{green}{$\uparrow$} & LLM-based Semantic Score 
& -
& LLM-judged semantic quality of generated text.

& \cite{cheng2025stola,cho2025ra,yu2025demonstrating} \\

MC & \textcolor{green}{$\uparrow$} & Material Consistency 
& -
& Agreement of material predictions across generated and conditioning data.

& \cite{fang2024bidirectional} \\

METEOR & \textcolor{green}{$\uparrow$} & METEOR Score 
& -
& Text similarity with synonym and stem matching.

& \cite{cheng2025stola} \\

\hline
\rowcolor{purple!10}
\multicolumn{6}{l}{\textbf{Metrics for Multimodal Manipulation and Control}} \\
\hline

SR & \textcolor{green}{$\uparrow$} & Success Rate 
& $\frac{N_{\text{success}}}{N_{\text{trials}}}$
& Ratio of successful executions.

& \cite{calandra2018more,hansen2022visuotactile,fazeli2019see,lee2020making,hao2025tla,zhang2025vtla,heng2025vitacformer,mao2024multimodal,zhu2025touch,feng2025anytouch,chen2022visuo} \\

IoU & \textcolor{green}{$\uparrow$} & Intersection over Union 
& $\frac{|G_p \cap G_g|}{|G_p \cup G_g|}$
& Overlap between predicted and ground-truth grasp regions.

& \cite{tong2025can} \\

AE & \textcolor{red}{$\downarrow$} & Angle Error 
& $|A_p-A_g|$
& Angular error of predicted grasp orientation.

& \cite{tong2025can} \\

GCR & \textcolor{green}{$\uparrow$} & Goal Convergence Rate 
& 
{\parbox[t]{4.2cm}{$\begin{aligned}
\frac{1}{m}\sum_{i=1}^{m}\mathbb{I}\!\big(&|\hat{x}_i-x_i|<\epsilon_x,\ |\hat{y}_i-y_i|<\epsilon_y,\\
&|\hat{r}_{z,i}-r_{z,i}|<\epsilon_r\big)
\end{aligned}$}}
& Fraction of actions within target tolerances.

& \cite{hao2025tla,zhang2025vtla,hansen2022visuotactile} \\

L1 & \textcolor{red}{$\downarrow$} & L1 Distance 
& $\frac{1}{m}\sum_{i=1}^{m}\|\mathbf{a}_i-\hat{\mathbf{a}}_i\|_1$
& Mean absolute action error.

& \cite{zhang2025vtla,hao2025tla} \\

HNS & \textcolor{green}{$\uparrow$} & Human Normalized Score 
& $\frac{\sum_{i=1}^{N}w_i s_i}{3\sum_{i=1}^{N}w_i}$
& Stage-weighted normalized manipulation score.

& \cite{heng2025vitacformer} \\

AP & \textcolor{green}{$\uparrow$} & Average Precision 
& -
& Precision-recall area for ranked predictions.

& \cite{li2026vtlg} \\

mAP & \textcolor{green}{$\uparrow$} & Mean Average Precision 
& $\frac{1}{C}\sum_{c=1}^{C}\mathrm{AP}_c$
& Mean AP across tasks, objects, or instances.

& \cite{li2026vtlg} \\

\bottomrule
\end{tabular}}
\end{table*}

\subsection{Evaluation Metrics}
\label{subsec:eval_metrics}

Evaluation metrics quantify model performance under a given benchmark setting. Unlike benchmarks, which define datasets and experimental protocols, metrics specify how prediction accuracy, generation quality, or control effectiveness is measured. Because multimodal tactile fusion spans diverse tasks and sensing modalities, no single metric is suitable for all scenarios. Instead, different task categories rely on distinct metrics according to their objectives. Table~\ref{tab:metrics_summary} summarizes representative metrics used in this field.

\subsubsection{Metrics for Multimodal Perception and Recognition}
\label{subsubsec:metrics_perception}

Multimodal perception and recognition constitute a major evaluation category in multimodal tactile fusion, including object recognition, attribute and material recognition, grasp success/failure prediction, and cross-modal retrieval and matching. Although these tasks differ in supervision and objectives, their evaluation metrics can generally be grouped into classification-based, similarity-based, and ranking-based measures.

\paragraph{Multimodal Object Recognition}
Multimodal object recognition is usually formulated as a supervised multi-class classification task, in which tactile and vision features are jointly used to predict object categories. Common metrics include precision~\cite{li2023vito,chen2025object}, recall~\cite{li2023vito,chen2025object}, accuracy~\cite{luo2018vitac,li2023vito,qiu2024dt,zheng2020lifelong,calandra2017feeling,wang2025damf,chen2025object,feng2025anytouch}, and F1-score~\cite{li2023vito,li2025tvt,wang2025damf,qiu2024dt,chen2025object}. These metrics directly evaluate classification performance, while F1-score is particularly informative under class imbalance.

\paragraph{Multimodal Attribute and Material Recognition}
Attribute and material recognition aims to infer physical properties such as material type or hardness. Depending on the task formulation, these problems may be cast as multi-class or binary classification tasks. Accordingly, accuracy and F1-score are commonly reported, and macro-averaged variants are often preferred under imbalanced class distributions.

Recent studies increasingly evaluate attribute and material recognition in zero-shot settings, where no task-specific labels are available during training. In this case, predictions are obtained by comparing tactile embeddings with text embeddings of candidate attribute descriptions, typically using cosine similarity~\cite{fu2024touch,yang2024binding,ma2025cltp,wu2025convitac,dave2024multimodal}. The predicted class is then selected according to the highest similarity score. This setting emphasizes semantic alignment and cross-task generalization rather than conventional supervised classification.

\paragraph{Grasp Success/Failure Prediction}
Grasp success/failure prediction determines whether a robotic grasp will succeed based on tactile observations before or during contact. It is typically formulated as a binary classification task and evaluated using accuracy, precision, recall, and F1-score. Because successful and failed grasps are often imbalanced, F1-score and recall for failure cases are frequently emphasized in safety-critical settings.

\paragraph{Cross-Modal Retrieval and Matching}
Cross-modal retrieval and matching evaluate whether tactile signals can be aligned with corresponding vision or language representations. These tasks are usually assessed with ranking-based metrics, particularly R@$k$, mAP, and MRR~\cite{ma2025cltp,dave2024multimodal,yang2024binding,cheng2025touch100k,cho2025ra,cheng2024towards}. Unlike classification metrics, these measures focus on cross-modal alignment quality and are therefore well suited for evaluating transferable tactile representations, including zero-shot and cross-sensory retrieval.

\subsubsection{Metrics for Multimodal Cross-Modal Generation and Transformation}
\label{subsubsec:metrics_generation}

Multimodal cross-modal generation and transformation synthesize one modality from another, such as generating tactile signals from vision or language inputs, or producing vision or language outputs from tactile observations. Evaluating these tasks is challenging because generation quality depends not only on low-level reconstruction fidelity, but also on perceptual realism, semantic consistency, and task relevance. Existing metrics can therefore be grouped into signal-level similarity, perceptual and distribution-based quality, semantic consistency, and LLM-based evaluation.

\paragraph{Vision-Tactile Generation and Translation}
For vision-tactile generation and tactile-guided image synthesis, evaluation usually begins with signal-level reconstruction metrics, including MSE and PSNR~\cite{fang2024bidirectional,li2019connecting}. Because these metrics mainly reflect reconstruction fidelity and may not correlate well with perceptual quality, prior work also reports SSIM and FID~\cite{fang2024bidirectional} to assess structural similarity and distribution-level realism. Beyond generic generation quality, some studies further adopt task-driven metrics such as classifier-based accuracy~\cite{fang2024bidirectional,yang2022touch,fu2024touch,xie2025universal} to examine whether generated samples preserve object- or material-level semantics. More recent work also uses embedding-based consistency scores derived from contrastive vision-tactile representations~\cite{yang2024binding,ma2025cltp} to measure whether generated tactile outputs remain aligned with their conditioning inputs in a shared embedding space.

\paragraph{Language-Tactile Generation and Translation}
For language-tactile generation, evaluation places greater emphasis on semantic consistency than on direct signal reconstruction. In tactile-to-language generation, recent work increasingly uses LLMs as automatic judges~\cite{cheng2025stola,cho2025ra,yu2025demonstrating}, assigning similarity or quality scores to evaluate semantic correctness and instruction adherence. This strategy provides a scalable proxy for human judgment. For text-to-tactile generation, direct sample-wise comparison is often less informative, so semantic consistency is typically evaluated indirectly through downstream tasks such as material classification consistency~\cite{fang2024bidirectional}. In this setting, generated tactile samples are fed into a pre-trained classifier, and evaluation examines whether the predicted labels match the semantics implied by the conditioning text or paired vision input. This protocol indicates whether the generated tactile outputs preserve discriminative physical properties relevant to downstream perception.

\subsubsection{Metrics for Multimodal Interaction and Manipulation}
\label{subsubsec:metrics_manipulation}

Multimodal interaction and manipulation assess whether tactile feedback improves contact-rich decision-making and action execution. Compared with perception tasks, their metrics are more closely tied to physical outcomes and therefore vary substantially across task settings. Existing metrics can be grouped into grasp detection, goal-oriented control, fine manipulation, dexterous manipulation, and ranking-based evaluation for language-conditioned manipulation.

\paragraph{Stable Grasping and Grasp Detection}
For stable grasping and grasp pose estimation, evaluation is often adapted from planar grasp detection. A predicted grasp is typically regarded as correct when both its orientation and spatial overlap with the ground truth satisfy predefined thresholds, which are commonly measured by angle error and IoU~\cite{tong2025can}. These metrics directly assess geometric grasp quality and are widely used in tactile-assisted grasp detection.

\paragraph{Goal-Oriented Tactile Control and Reaching}
For tactile-guided reaching and contact, evaluation mainly focuses on whether the target contact is successfully achieved. The most common metric is the binary success rate~\cite{calandra2018more}, which directly reflects whether multimodal feedback enables accurate contact localization under partial observability.

\paragraph{Fine Manipulation and Assembly Tasks}
For fine manipulation and assembly, which require continuous control over multiple dimensions, evaluation often measures both convergence and action accuracy. Common metrics include Goal Convergence Rate (GCR)~\cite{hao2025tla,zhang2025vtla,hansen2022visuotactile} and L1 distance~\cite{zhang2025vtla,hao2025tla}. Compared with binary success rate, these metrics provide more fine-grained information about control precision and whether predicted actions remain within task-specific tolerances.

\paragraph{Dexterous Manipulation Metrics}
For dexterous manipulation, simple success rates often fail to reflect intermediate progress and the specific contribution of tactile feedback. Stage-aware metrics such as Human Normalized Score (HNS)~\cite{heng2025vitacformer} therefore provide a more detailed evaluation by accounting for both task progression and tactile dependence.

\paragraph{Language-Conditioned Multimodal Manipulation}
For language-conditioned manipulation, evaluation often uses ranking-based metrics similar to those used in retrieval and detection. Given multiple predicted grasps or actions, Precision@$k$, AP, and mAP~\cite{li2026vtlg} are commonly reported to evaluate whether language-guided policies produce correct and task-relevant actions across diverse instructions.

\subsubsection{Limitations of Existing Metrics}
\label{subsubsec:metrics_limitations}

Despite their diversity, existing metrics remain fragmented and highly task-specific. Classification and retrieval metrics reflect recognition and alignment quality, but they do not capture physical feasibility. Generation metrics measure reconstruction fidelity or semantic consistency, yet often correlate weakly with tactile realism and downstream utility. Manipulation metrics better reflect task execution, but they are usually tied to specific embodiments, sensors, and action spaces, which limits comparability across studies. As a result, no single metric currently captures tactile realism, perceptual plausibility, semantic correctness, and control effectiveness in a unified manner.

\subsection{Benchmarks}

Unlike more established areas such as vision-language learning, multimodal tactile fusion still lacks unified benchmarks. Existing evaluation settings are instead largely task-driven and dataset-specific, with recurring combinations of datasets, task formulations, and experimental protocols serving as de facto benchmarks.

\subsubsection{Benchmarks for Multimodal Perception and Recognition}

Benchmarks for multimodal perception and recognition mainly evaluate whether tactile signals improve recognition, attribute inference, grasp outcome prediction, and cross-modal retrieval when combined with modalities such as vision or proprioception. Typical settings are built on representative vision-tactile datasets such as VisGel~\cite{li2019connecting}, ViTac~\cite{luo2018vitac}, UniTouch~\cite{yang2024binding}, and Touch and Go~\cite{yang2022touch}, usually under fixed train-test splits, closed-set protocols, or shared-embedding retrieval settings.

\subsubsection{Benchmarks for Multimodal Cross-Modal Generation and Transformation}

Benchmarks for multimodal cross-modal generation and transformation evaluate whether a model can translate or synthesize one modality from another, such as generating tactile signals from images or producing language descriptions from tactile input. Common settings are built on paired multimodal datasets such as VisGel~\cite{li2019connecting}, UniTouch~\cite{yang2024binding}, TLV-Link~\cite{cheng2025touch100k}, and Touch in the Wild~\cite{zhu2025touch}, with evaluation typically combining reconstruction, semantic consistency, retrieval, and model- or human-based protocols.

\subsubsection{Benchmarks for Multimodal Interaction and Manipulation}

Benchmarks for multimodal interaction and manipulation evaluate how tactile information supports decision-making and closed-loop action execution in embodied systems. They are commonly built on robotic manipulation datasets or platforms such as More than a Feeling~\cite{calandra2018more}, AnyTouch~\cite{feng2025anytouch}, VTLA~\cite{zhang2025vtla}, and VTLG~\cite{li2026vtlg}, with representative tasks including grasping, contact-rich manipulation, and language-guided control, typically measured by task-level success or action quality.

\subsection{Cross-Task and Cross-Dataset Comparison}

A central question in multimodal tactile fusion is whether learned representations can generalize across tasks and datasets. Because existing models are evaluated under different task definitions, sensing setups, and data distributions, such cross-task and cross-dataset comparisons are important for understanding robustness and transferability.

\subsubsection{Generalization across Tasks}

Early multimodal tactile methods were usually designed for individual tasks, such as object recognition or grasp prediction, and their learned representations often transferred poorly beyond the original setting. More recent work instead aims to learn more general multimodal representations. For example, Touch and Go~\cite{yang2022touch} and UniTouch~\cite{yang2024binding} show that aligning tactile signals with vision or language can improve transfer across tasks with limited additional supervision. This trend suggests that cross-task generalization increasingly depends on representation quality rather than task-specific design.

\subsubsection{Generalization across Datasets}

Cross-dataset generalization remains more challenging because tactile observations are strongly influenced by sensor hardware, contact conditions, and data collection protocols. As a result, models often exhibit substantial sensor bias and dataset bias. Recent studies suggest that alignment-based methods and unified multi-sensor learning can improve transfer, but performance still degrades noticeably when sensing setups differ substantially. Current evaluations therefore remain largely in-distribution.

\subsubsection{Effects of Data Scale, Modality Coverage, and Pretraining}

Generalization is also strongly influenced by data scale, modality coverage, and pretraining. Because tactile data are expensive to collect, small datasets often lead to overfitting and limited transfer. Larger multimodal datasets and richer modality combinations, especially those involving language, generally improve representation quality. Pretraining is particularly helpful because it allows tactile models to benefit from strong priors in vision and language encoders, improving both benchmark performance and cross-task transfer.
\section{Discussion}
\label{sec:discussion}

Although multimodal tactile fusion has made substantial progress, several  bottlenecks still limit its development from task-specific systems to more general embodied intelligence. Therefore, we summarize several current challenges that reflect both practical difficulties and broader research opportunities.

\subsection{Current Challenges}

\textbf{(1) Data Scalability Gaps.}
Existing datasets (Section~\ref{sec:datasets}) are often task-specific, sensor-dependent, and much smaller than large-scale vision-language resources. This gap creates a mismatch with data-intensive foundation models, limiting the physical grounding of language semantics and reducing zero-shot transfer capability in embodied systems.
\textbf{(2) Modality Misalignment and Noise.}
Multimodal fusion methods still struggle with spatiotemporal misalignment between sparse tactile inputs and dense vision or language inputs. In unstructured settings, sensor drift, visual occlusion, and missing observations can further weaken cross-modal alignment, reducing the reliability of generation and increasing the risk of manipulation failure.
\textbf{(3) Hardware-Software Integration.}
The morphological diversity of tactile sensors, such as rigid and flexible designs, together with the lack of standardized interfaces, makes unified system integration difficult. In addition, limited durability and high power consumption restrict real-time closed-loop fusion with large vision-language models, creating a system-level bottleneck for whole-body embodied sensing.
\textbf{(4) Benchmark Deficiencies.}
Evaluation metrics (Section~\ref{sec:evaluation}) remain highly task-specific and are often separated across recognition, generation, and manipulation settings. The lack of holistic end-to-end embodied benchmarks limits cross-study comparison and makes it difficult to assess practical factors such as safety and robustness during physical interaction.

\subsection{Future Directions}

\textbf{(1) Scalable Multimodal Datasets.}
To break the data bottleneck of foundation models, future collection must prioritize temporally aligned, action-conditioned visuo-tactile trajectories over static pairs. Furthermore, leveraging physics-informed generative models (e.g., diffusion-based tactile hallucination) for large-scale data augmentation will be pivotal in overcoming physical scarcity and bridging the sim-to-real gap for zero-shot transfer.
\textbf{(2) Hierarchical Fusion Architectures.}
Future models should move beyond flat fusion toward hierarchical architectures that use tactile feedback as a grounded layer for multimodal reasoning. Self-supervised pretraining on physical interaction data will enhance robustness to noise and missing observations.
\textbf{(3) Bio-Inspired Sensors and Aquatic Environments.}
On the hardware side, tactile systems are expected to develop toward compliant, durable, and large-area e-skins with improved on-sensor processing. Future work should also extend multimodal tactile sensing to underwater and other visually degraded environments through waterproof and hybrid sensor designs.
\textbf{(4) End-to-End Systems and Safety.}
The paradigm must shift toward closed-loop embodied systems where tactile feedback acts as a continuous supervisory signal. To transition from controlled labs to unstructured human environments, explicitly embedding uncertainty estimation and physical compliance constraints directly into the policy generation loop is imperative.
\textbf{(5) Simulation Synergies and Generalization.}
The morphological diversity of tactile sensors remains a fundamental barrier. Future research must focus on unified tactile tokenization strategies and high-fidelity, physics-based contact simulations to facilitate cross-domain adaptation. This synergy will enable meta-learning frameworks capable of seamlessly transferring manipulation skills across heterogeneous robotic embodiments.
\section{Conclusion}
\label{sec:Conclusion}
This survey comprehensively reviews multimodal tactile fusion across datasets, hardware, learning paradigms, and evaluation. Our primary contribution is conceptualizing a unified physical-to-computational pipeline that explicitly bridges raw sensor transduction with high-level semantic reasoning. While the field is rapidly shifting toward generalized VLA foundation models, critical bottlenecks, specifically hardware fragmentation, the scarcity of action-conditioned data, and the sim-to-real gap, continue to impede the true physical grounding of embodied agents.
Looking forward, as systems venture into unpredictable and visually degraded domains, tactile fusion will transcend its role as a mere supplementary modality. Resolving current modality misalignments and integration barriers will establish tactile feedback as the indispensable physical cornerstone of embodied AI, ultimately empowering autonomous agents with the robust, contact-aware intelligence required for general-purpose execution in the wild.

\ifCLASSOPTIONcaptionsoff
  \newpage
\fi



\small
\bibliographystyle{IEEEtran}
\bibliography{ref}
\end{document}